\documentclass[10pt]{article} 
\usepackage[preprint]{tmlr}

\usepackage{hyperref}
\usepackage{url}

\usepackage{hyperref}
\usepackage{url}
\usepackage{graphicx} 
\usepackage{booktabs}
\usepackage{colortbl}
\usepackage{amssymb}
\usepackage{bbding}
\usepackage{pifont}
\usepackage{amsthm}
\usepackage{chngcntr}
\counterwithout{table}{section}
\usepackage{amsmath}

\newtheorem{theorem}{Theorem}

\newtheorem{proposition}[theorem]{Proposition}

\newtheorem{property}{Property}

\theoremstyle{remark}
\newtheorem{remark}{Remark}

\theoremstyle{definition}

\title{Latent Space Disentanglement in Diffusion Transformers Enables Zero-shot Fine-grained Semantic Editing}

\author{Zitao Shuai, Chenwei Wu\thanks{co-first author}, Zhengxu Tang\thanks{co-first author}, Bowen Song \& Liyue Shen \thanks{coresponding author} \\
University of Michigan\\
\texttt{\{liyues\}@umich.edu}
}

\def\month{MM}  
\def\year{YYYY} 
\def\openreview{\url{https://openreview.net/forum?id=XXXX}} 

\begin{document}

\maketitle

\begin{abstract}
Diffusion Transformers (DiTs) have achieved remarkable success in diverse and high-quality text-to-image(T2I) generation. However, how text and image latents individually and jointly contribute to the semantics of generated images, remain largely unexplored. Through our investigation of DiT's latent space, we have uncovered key findings that unlock the potential for zero-shot fine-grained semantic editing: (1) Both the text and image spaces in DiTs are inherently decomposable. (2) These spaces collectively form a disentangled semantic representation space, enabling precise and fine-grained semantic control. (3) Effective image editing requires the combined use of both text and image latent spaces. Leveraging these insights, we propose a simple and effective Extract-Manipulate-Sample (EMS) framework for zero-shot fine-grained image editing. Our approach first utilizes a multi-modal Large Language Model to convert input images and editing targets into text descriptions. We then linearly manipulate text embeddings based on the desired editing degree and employ constrained score distillation sampling to manipulate image embeddings. We quantify the disentanglement degree of the latent space of diffusion models by proposing a new metric. To evaluate fine-grained editing performance, we introduce a comprehensive benchmark incorporating both human annotations, manual evaluation, and automatic metrics. We have conducted extensive experimental results and in-depth analysis to thoroughly uncover the semantic disentanglement properties of the diffusion transformer, as well as the effectiveness of our proposed method. Our annotated benchmark dataset is publicly available at https://anonymous.com/anonymous/EMS-Benchmark, facilitating reproducible research in this domain.
\end{abstract}

\section{Introduction}
\begin{figure*}[t!] 
    \centering
    
    \includegraphics[width=1\textwidth]{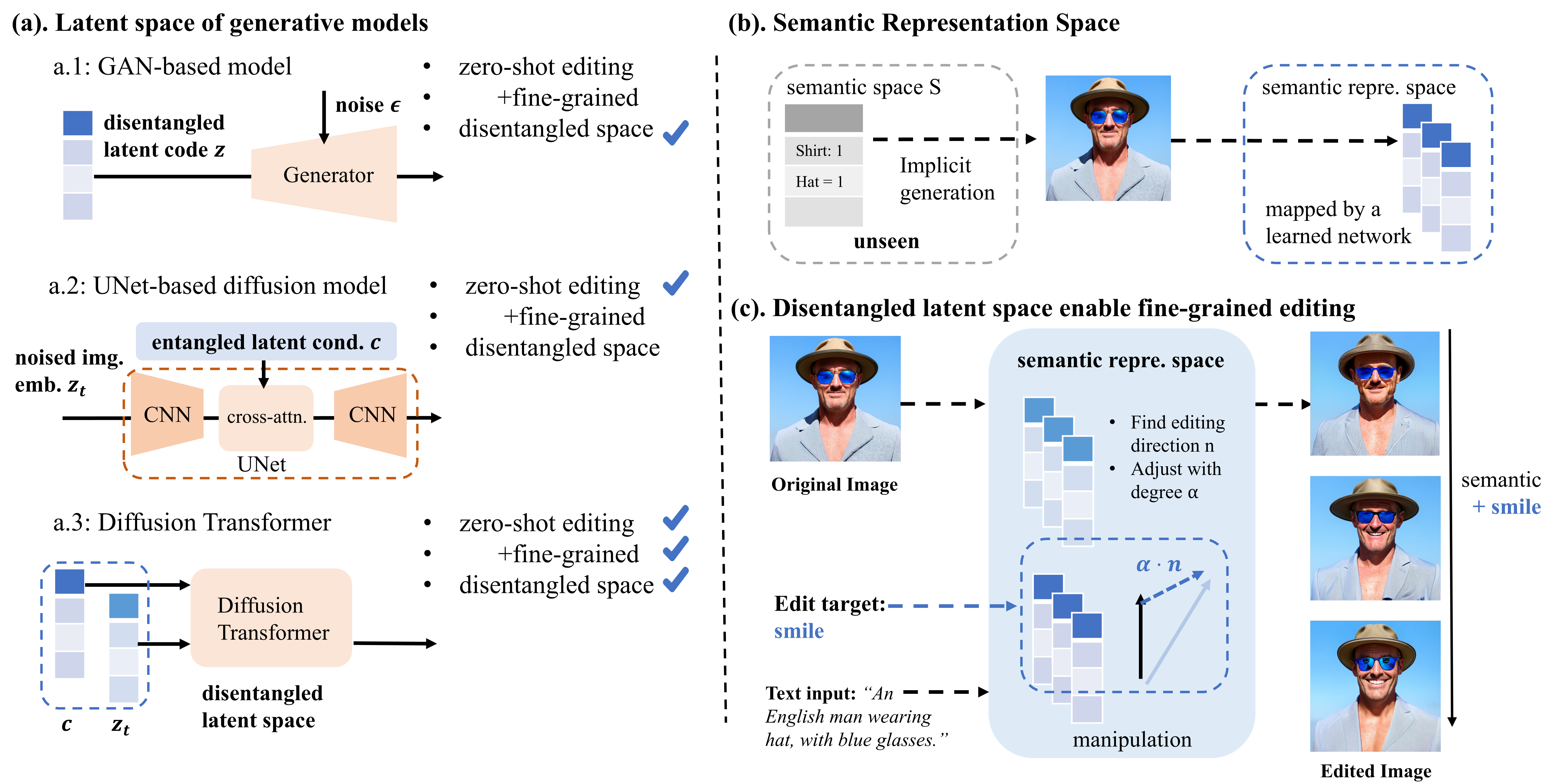} 
    \caption{
    (a). Comparisons of latent spaces of GAN-based models, UNet-based diffusion models, and diffusion transformers. (b). In classic modeling on visual generation and feature representation~\cite{wang2021self}, images are generated by unseen semantics through implicit mapping. We aim to find a semantic representation space learned by neural networks to manipulate visual semantics. (c). Text-to-image diffusion transformer possesses a disentangled semantic representation space, which facilitates precise and fine-grained editing on target semantics. We achieve this by identifying disentangled editing directions and modifying the original representations within the semantic representation space.
    }
    \label{figure1}
\end{figure*}

Recently, diffusion transformers~\cite{esser2024scaling, peebles2023scalable} have proven to be very effective in synthesizing diverse and high-fidelity samples following text conditions. Unlike UNet-based T2I diffusion models~\cite{rombach2022high}, where a fixed text representation is fed directly into the model via cross-attention, the denoising blocks of diffusion transformers intake text and noised image representation sequences as a whole, enabling a two-way flow of information, as shown in Fig.~\ref{figure1} (a). While empirical results have shown that T2I generation benefits from this new structure of image and text latent space~\cite{esser2024scaling}, how they (individually and jointly) contribute to generated image semantics remains unexplored.

Previous works~\cite{shen2020interpreting,shen2021closed} on pre-diffusion generative models, such as StyleGAN~\cite{karras2019style}, have shown that the ability to generate controllable outputs comes from the semantic disentanglement properties of their latent representation spaces. For example, in image editing, the disentangled latent space of StyleGAN allows for the editing of facial semantics with fine granularities. Given a facial image, we first obtain its latent code and then identify an editing direction corresponding to the "smile" semantic. By adjusting the latent embedding along this direction, we can control the intensity of the smile reflected in the edited image. Although~\cite{hertz2022prompt,brooks2023instructpix2pix} have demonstrated that UNet-based diffusion models possess inherent word-to-semantic mappings, similar semantic disentanglement effects have not been observed in these models, and thus fine-grained controlled image synthesis remains a challenge~\cite{lu2024hierarchical,kwon2022diffusion}. Therefore, we aim to explore whether the diffusion transformer exhibits semantic disentanglement effects within its latent space and, if so, how to leverage them to conduct image synthesis tasks with fine-grained control.  

Motivated by this, we take the first step in uncovering the semantic disentanglement in the latent space of T2I transformer-based diffusion models. We start by hypothesizing and modeling the disentanglement property of the T2I diffusion transformer's latent space. (1) We verified the decomposability in representation spaces constructed by text and image embeddings, which allows for linear manipulation to modify specific semantics with fine granularity without affecting others as demonstrated in Fig~\ref{figure1} (c). (2) We found that the text embeddings and noised image embeddings collectively form a disentangled semantic representation space. (3) Effective image editing requires the combined use of both text and image latent spaces.

Based on these findings, we propose an \textbf{E}xtract-\textbf{M}anipulate-\textbf{S}ample (EMS) method for zero-shot precise and fine-grained editing with a T2I diffusion transformer. Specifically, to leave non-editing targets unchanged, we extract the semantics of the given image as well as that of the desired image into text prompts with a multi-modal Large Language Model. To adjust the degree of the semantics move toward desired values, we linearly manipulate embeddings of both original and target text prompts in the latent space. For image-side manipulations, we propose a constrained score-distillation-sampling method to manipulate image latent. We additionally propose a \textbf{S}emantic \textbf{D}isentanglement m\textbf{E}tric (SDE) to measure the disentanglement property of the latent spaces of T2I diffusion models. In the absence of comprehensive fine-grained image-editing evaluation datasets, we construct ZOFIE, a novel \textbf{Z}ero-shot \textbf{O}pen-source \textbf{F}ine-grained \textbf{I}mage \textbf{E}diting benchmark. Both automatic and human evaluations demonstrate the effectiveness of our framework on fine-grained editing tasks.

Our contributions are summarized as follows:
\begin{itemize}
    \item We are the first to investigate the existence and origins of DiT's semantic disentanglement properties via quantitative and qualitative analysis, validating the feasibility of linear manipulation on both image and text embeddings for precise and fine-grained control. 
    \item We propose the EMS method for zero-shot disentangled and fine-grained editing. Our method is simple and efficient in precisely editing images without any additional training.
    \item We provide a benchmark with both automatic and human annotations to assess the effectiveness of disentangled and fine-grained editing tasks, as well as the disentanglement property of generative models.
    \item We propose a novel metric to quantify the disentanglement degree of latent spaces of T2I diffusion models.
\end{itemize}

\section{Exploration on Latent Space of Diffusion Transformer}
\label{exploration}
In this section, we first provide preliminaries of Text-to-Image (T2I) diffusion transformers, image editing. Then we will present our modeling of disentangled semantic representation space, and our findings on the semantic disentanglement properties of the latent space of the T2I diffusion transformer

\subsection{Preliminary}
\label{Preliminary}
\textbf{Text-to-Image Diffusion Transformer.} The main difference between network structures of Text-to-Image (T2I) diffusion transformer~\cite{esser2024scaling} and UNet-based T2I diffusion models, is that the former utilizes a transformer as the denoising block, building upon the DiT~\cite{peebles2023scalable} architecture. It employs an in-context conditioning approach to integrate both text and image modalities for text-conditional image sampling. 
T2I diffusion transformer also has an VAE encoder to encode input image to latent space, an VAE decoder to transform image embeddings back to the pixel space, a de-noisng transformer block, and a set of text encoders to encode text prompts, 
providing semantics with different levels of granularity. 

\label{latent space}
During inference, the text prompt will be transformed into token-type text embeddings $ c \in \mathbb{R}^{l\times d} $, and pooled text embedding $ c_{\text{pool}} \in \mathbb{R}^{d}$ via averaging $c$, where $l,d\in  \mathbb{N}^+$. In the image editing task, given a source image, the vision encoder first encode it to latent representation and patchify it to token-type image embeddings $ z_0 \in \mathbb{R}^{v\times d} $, where $v\in  \mathbb{N}^+$. 
For forward and reverse process, ~\cite{esser2024scaling} considers both rectified flow matching sampling scheduler and traditional LDM~\cite{rombach2022high} framework. In this paper, we mainly focus on exploring the properties of learned latent space, and simplify these process as following. 
For de-noising loops, both the noised image tokens and text tokens will be input to the de-noising transformer. After that, the VAE decoder will decode image latents to pixel space. In the following sections, we mainly consider the image latent space composed of $z_0$, noised image latent space composed of $z_t$, and text latent space composed of $c$. 



\textbf{Image Editing}. Image editing aims to modify certain semantic contents in the given source image $x$, while leaving other semantics unchanged. In this paper, we primarily focus on text-guided image editing, as we often edit images based on text instructions in practical applications. Following~\cite{wu2023uncovering}, we model the image editing as the two following processes on the latent space. (1). Image forward Process: given encoded image latents $z_0$ and timestep $t$ which measures the noise degree, we add noise and then patchify it to get the image embedding $ z_t$. (2). Image Reverse Process: we input text prompts for editing to text encoders to gain $ \tilde{c_{\text{pool}}}, \tilde{c} $, then utilize them as the input condition of the de-noising transformer, and get a de-noised image latent $\tilde{z}_0$. Then we decode the $\tilde{z}_0$ to obtain the edited target image $\tilde{x}$. In this paper, we focus on precise and fine-grained editing on considered semantics with a frozen T2I diffusion transformer. Given a semantic $s$ and a source image $x$, \textit{we aim to control the degree $\alpha$ to which semantic $s$ moves toward target value in the edited target image $\tilde{x}$, and avoid incorrectly changing other semantics.} Our key areas of interest and differences to the most related works are summarized in Table \ref{table1}.

\begin{table}[h] 
\centering 

\caption{Comparisons between related image editing works with diffusion models and our considered precise and fine-grained zero-shot image editing via EMS. We organize the related works based on our key areas of interest: (1). Achieve fine-grained editing of a given semantic. (2). Zero-shot image editing using a frozen pre-trained model. (3). No need for an image or text as a reference in inference time. (4). Be able to precisely edit target semantics without changing others.}
\scalebox{0.77}{
\begin{tabular}{@{} l llllll }
\toprule
\textbf{Method} &\textbf{Description} & \textbf{Fine-grained}& \textbf{Zero-Shot} & \textbf{Reference-Free} & \textbf{Precise Modify} \\
\midrule
~\cite{hertz2022prompt}& prompt-to-prompt image editing & & $\checkmark$& & \\
~\cite{cao2023masactrl}& mutual self-attention for editing & & $\checkmark$& & \\
~\cite{brooks2023instructpix2pix}& tuning model with edit instruction& & &$\checkmark$ &$\checkmark$ \\
\midrule
~\cite{couairon2022diffedit}& mask semantic irrelevant part then edit & &$\checkmark$ & $\checkmark$ & \\
~\cite{hertz2023delta} & optimize image latents for editing & & $\checkmark$ & $\checkmark$ & $\checkmark$ \\
\midrule
~\cite{wu2023uncovering} & optimize text embedding for editing& & &$\checkmark$ & \\
~\cite{kawar2023imagic} & optimize text embedding for editing& $\checkmark$ & & $\checkmark$ & $\checkmark$ \\
\midrule
EMS (ours) & Extract-Manipulate-Sample& $\checkmark$ & $\checkmark$ & $\checkmark$ & $\checkmark$\\
\bottomrule
\end{tabular}
}
\label{table1}
\end{table}

\subsection{Disentangled Semantic Representation Space of Diffusion Transformer}
\label{disentangle space sec}
In this section, we will first introduce the definition and ideal properties of disentangled semantic representation space. Then, we will present our analysis and findings on whether such properties exist in the diffusion transformer's latent space.

\subsubsection{Definitions of Disentangled Semantic Representation Space and Key Properties}
\label{def disentangle}
In the classic modeling of visual generation and feature representation~\cite{wang2021self, higgins2018towards}, let $\mathcal{S}$ represents a set of unseen semantics (e.g., "color" and "style"), where each semantic can take on various values (e.g., color: "purple" or "red"). There exists an generation process $\Phi: \mathcal{S} \rightarrow \mathcal{I}$ that generates images from semantics (e.g., thinking about a red dress $\rightarrow$ painting a red dress). Ideally, $\mathcal{S}$ can be decomposed into the Cartesian product of $m$ modular semantic subspaces $\mathcal{S} = \mathcal{S}_1 \times \ldots \times \mathcal{S}_m, m \in \mathbb{N}^+$~\cite{yue2024exploring}. With this property, we can locally intervene the modular semantic $s_i \in \mathcal{S}_i, i \in [1,\ldots,m]$ without affecting the others (e.g., in facial images, changing the value of semantic "glasses" will not modify "ethnicity" and "expression"). Then we can consider all image editing operations (e.g., changing semantic "expression") as a group $G$ acting on $\mathcal{S}$, which can be decomposed as the direct product of them: $G = g_1 \times \ldots \times g_m$. The intervention on semantic $s_i$ can be defined as the operator $g_i: \mathcal{S}_i \rightarrow \mathcal{S}_i$, which manipulates $s_i$ by $\tilde{s_i} = g_i(s_i)$. 

However, $\mathcal{S}$ is an ideal, unobservable space and therefore cannot be directly leveraged for downstream generation or editing tasks. In visual representation learning, we typically approximate $\mathcal{S}$ by learning a semantic representation space $\mathcal{S}'$ through a neural network $f: \mathcal{I} \rightarrow \mathcal{S}'$.

In this work, for image editing with a diffusion transformer, we consider this semantic representation space $\mathcal{S}'$ is jointly formed by noised image embeddings and text embedding. As discussed in Sec.~\ref{Preliminary}, the input image is first encoded by a fixed VAE encoder into image embeddings $z_0$, which are then transformed into noised image embeddings $z_t$ during the forward process, forming the representation space $\mathcal{Z}_t$. Simultaneously, the text prompt is encoded into text embeddings $c$, constituting the representation space $\mathcal{C}$. In image generation, $z_t$ and $c$ are first denoised to $z_0$, and then decoded back to an image in $\mathcal{I}$. We define this process of denoising and decoding as function $h: \mathcal{S}' \rightarrow \mathcal{I}$.

In image editing, we intervene on target semantic $z_t$ or $c$, and obtain \( \tilde{z_t}, \tilde{c} \). We then employ \( h \) to map them from semantic representation space to get the edited image. However, the intervention on semantic representation space $\mathcal{S}'$ might be inaccurate and not controllable if $\mathcal{S}'$ is not disentangled. To achieve precise and fine-grained editing, $\mathcal{S}'$ needs to be a disentangled space that satisfies the following properties.

\begin{property}
\label{remark decompose}
\textbf{Decomposable:} For a decomposable semantic representation space $S'$, there exists a decomposition\footnote{Here in the decomposability, we have not assumed that semantics in training datasets or testing dataset are statistically independent. We focus on the property that "modifying desired semantic without changing others".} such that $S' = S'_1 \times \cdots \times S'_m$ corresponding to semantics in the real semantic space $S$. Therefore, $S'_i$ is only affected by $g_i$. For example, changing the "expression" vector in $S'$ does not affect the "ethnicity" semantic in the edited image.
\end{property}

\begin{property}
\label{remark effectiveness}
\textbf{Effectiveness:} For any editing operation $g_i \in \mathcal{G}$, and any pair of latent representation semantic and its corresponding real-world semantic $s_i \in \mathcal{S}$, there exists $h(g_i(s_i')) = \phi(g_i(s_i))$. For example, changing the latent representation semantic from "serious" to "smile" in the unseen $S'$ will be equivalent to directly changing the real image expression semantic in $\mathcal{S}$ with the direction from "serious" to "smile".
\end{property}

\begin{figure*}[t!] 
    \centering
    
    \includegraphics[width=1\textwidth]{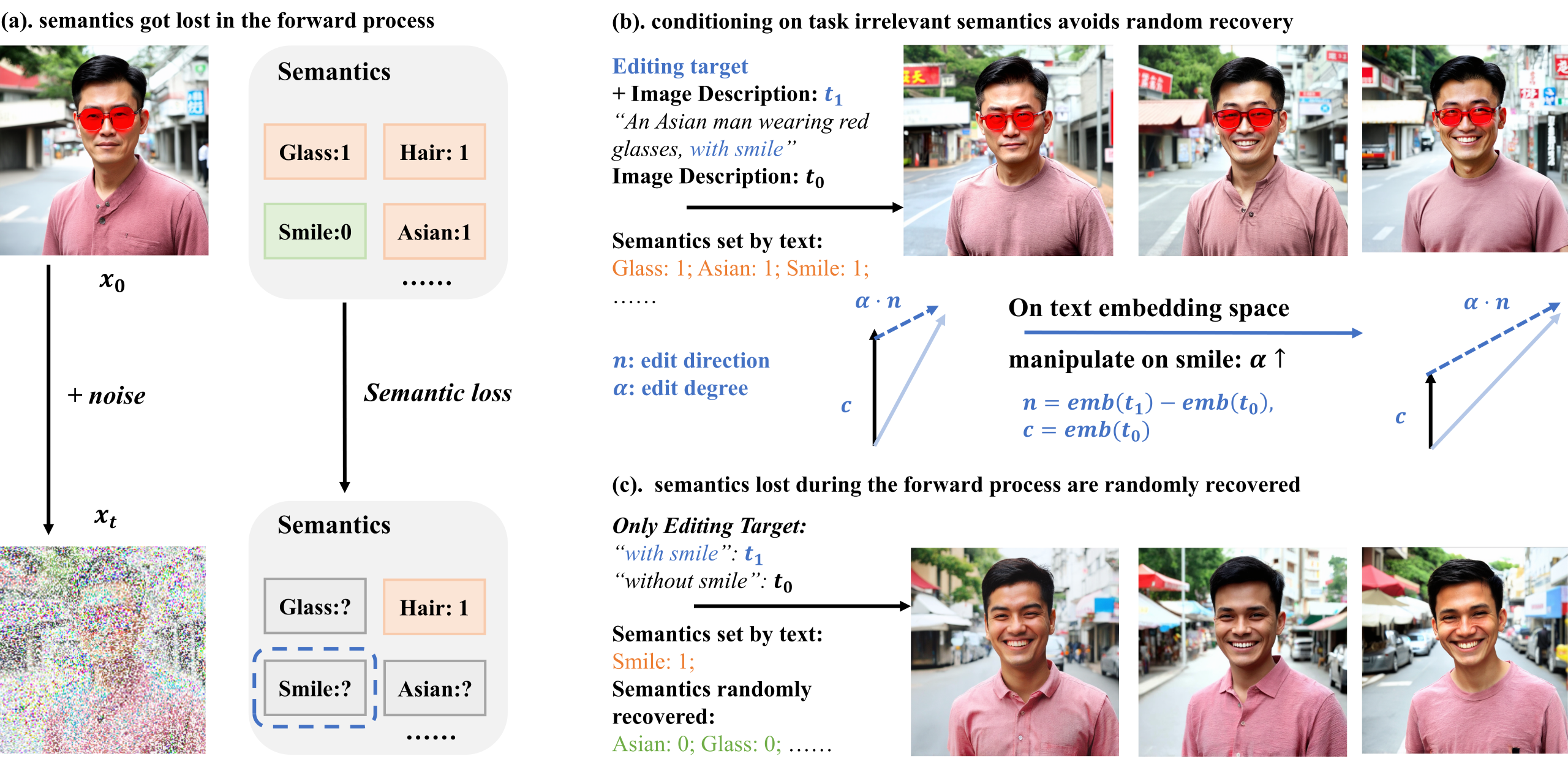} 
    \caption{(a). We observed the \textit{semantic loss} phenomenon, certain semantics will be lost during the forward process, and will be assigned random values in the reserve process if they are not conditioned. (b). The text latent space is decomposable, and provides explicit directions for adjusting how semantis move toward desired values in the edited images. For fine-grained and precise editing, we first obtain a text prompt of the source image, and gain a text prompt that describes the desired image. We then manipulate representations encoded from these two prompts. (c). Only conditioning on editing targets would result in inaccurate editing, where some task-irrelevant semantics are incorrectly modified.
    }
\label{figure2}
\end{figure*}

\subsubsection{Validation of Key Properties of Latent Space of Diffusion Transformer}
We have found that the semantic representation space $\mathcal{Z}_t \cup \mathcal{C}$ in the diffusion transformer is highly disentangled, as it satisfies the two key properties discussed above. In the following sections, we will present our theoretical and empirical verifications of this finding.

\label{text disentangle}
\textbf{The text embedding forms a decomposable semantic representation space $\mathcal{C}$, where we can obtain editing directions for modifying semantics.} We have successfully achieved precise modification on target semantics with text-side manipulation. Specifically, we first obtain text description of the source image, and encode it to get text embedding $c$. We then manipulate it by $\tilde{c}=c+\alpha \cdot n$, with a given editing degree $\alpha$ and editing direction $n$. As shown in Fig.~\ref{figure2}, this approach successfully modified the expression semantic to "smile" with fine granularity, without affecting other semantics. The editing direction $n$ can be determined explicitly, either by leveraging the text embedding corresponding to the desired semantic from a text prompt composed of relevant words or by subtracting the embeddings of two text prompts: $n = \text{emb}(t_1) - \text{emb}(t_0)$, where $t_0$ describes the source image and $t_1$ describes the edited image\footnote{In our method, we instruct the multi-modal LLM to generate text descriptions for both the original and desired images, ensuring that they differ only in the tokens corresponding to the semantics to be edited. Consequently, in the encoded text embeddings, only parts related to the specified semantics are altered.}. These empirical results highlight the decomposability of the text embedding space, enabling fine-grained editing capabilities.


\begin{figure*}[t!] 
    
    \centering
    \includegraphics[width=1\textwidth]{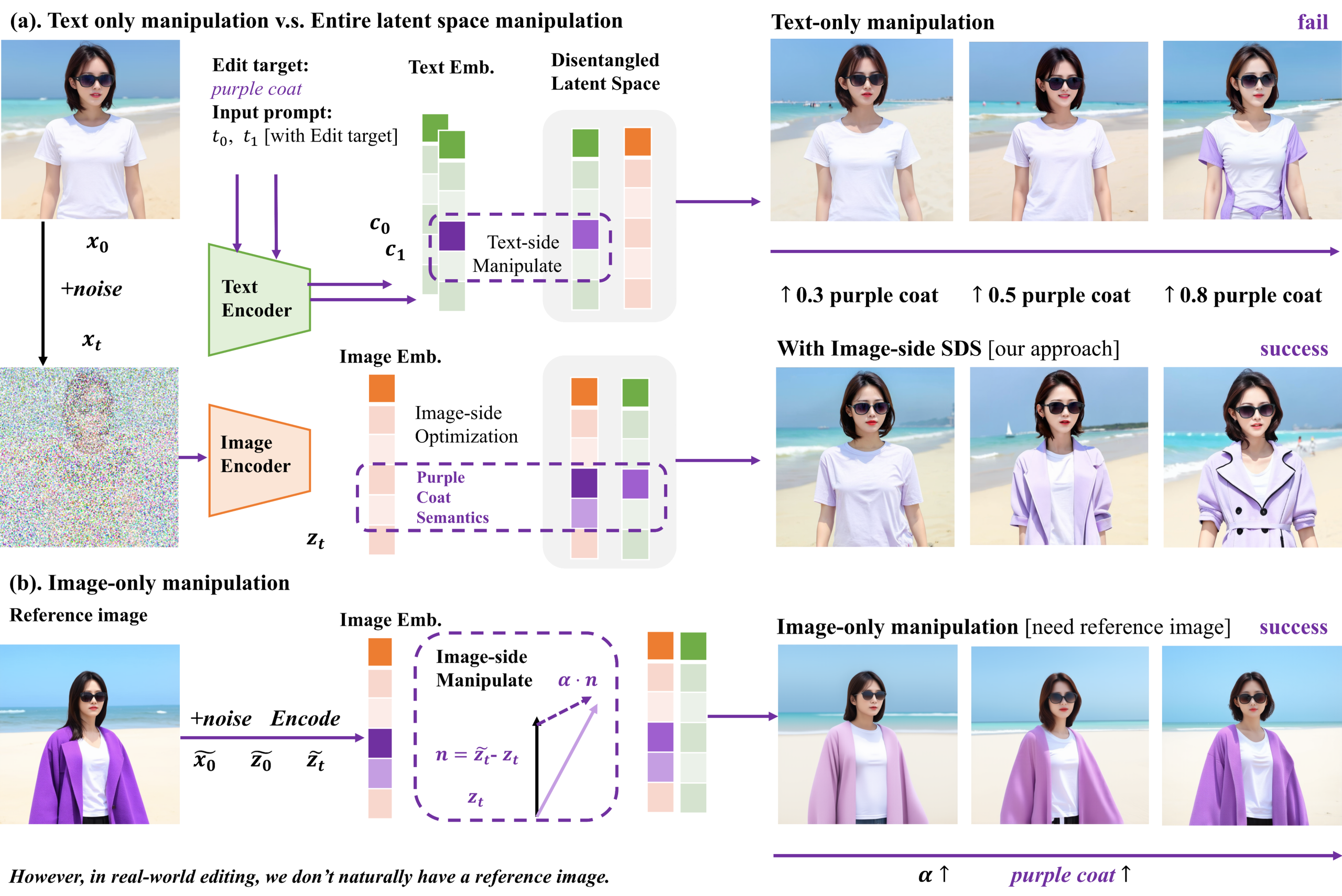} 
    \caption{(a). We observe that only manipulating text embeddings cannot effectively modify certain semantics. On the other hand, modifying the entire latent space, including score-distillation-sampling-based (SDS) manipulation on image embeddings, can help address this issue. (b). Given a reference image, image embeddings can also be linearly manipulated to achieve fine-grained editing. However, in real-world scenarios, an ideal reference image is often unavailable. 
    }
    \label{figure3}
\end{figure*}

\label{image disentangle}
\textbf{The latent space $\mathcal{Z}_t$ formed by image embedding can be manipulated to achieve fine-grained editing.} We have also observed the decomposability on the image side as shown in Fig.~\ref{figure3} (b). We select two images that differ only in the value of the semantic attribute "cover". We then encode them to two image embeddings $z_t$ and $\tilde{z}_t$ with the frozen diffusion transformer. We linearly manipulate them to get $z_{t, \alpha}= \alpha \cdot (\tilde{z}_t - z_t) +  z_t$, where the image embedding of the image with "purple coat" $z_t$ shifted toward desired $\tilde{z}_t$ different ratio $\alpha$. As we increase $\alpha$, the purple coat (of the cover semantic) gets more significant. Therefore, the image embedding space is also disentangled and can be linearly manipulated. Here we construct the editing direction with $\tilde{z}_t - z_t$. Therefore, similar to $\mathcal{C}$, $\mathcal{Z}_t$ can be represented by cartesian product of semantic subspaces: $\mathcal{Z}_t = Z_{t,1}\times \cdots \times Z_{t,m}, i\in[1,...,m]$.

\textbf{$\mathcal{Z}_t \cup \mathcal{C}$ satisfies both properties and forms a disentangled semantic representation space. However, neither $\mathcal{C}$ nor $\mathcal{Z}_t$ alone can achieve the effectiveness property.} We have verified the decomposability property of both $\mathcal{C}$ and $\mathcal{Z}_t$, thus their joint space $\mathcal{S}' = \mathcal{C} \oplus \mathcal{Z}_t$ is also decomposable. The effectiveness of $\mathcal{Z}_t \cup \mathcal{C}$ has been empirically verified in the experiments section. Therefore, $\mathcal{Z}_t \cup \mathcal{C}$ is a disentangled semantic representation space. 

However, neither $\mathcal{C}$ nor $\mathcal{Z}_t$ alone can achieve effectiveness in modifying semantics. Firstly, while $\mathcal{Z}_0$ captures all semantics, $\mathcal{Z}_t$ may lose some of them during the forward process. We have observed the semantic loss phenomenon, as demonstrated in Fig.~\ref{figure2} (a). This phenomenon describes that some attributes become noised during the forward process and are assigned random values during the reverse diffusion process from $z_t$ to the reconstructed $\hat{z}_0$ when the reverse process is not conditioned by labels or text embeddings. In Fig.~\ref{figure2} (c), when the semantics `ethnicity' and `glasses' are left unconditioned on the text side, the reconstructed samples exhibit random race and glasses attributes. In contrast, when these semantics are conditioned by text embeddings, the attributes are correctly recovered as shown in Fig.~\ref{figure2} (b). This observation indicates that some semantics cannot be effectively conditioned if text embeddings are not leveraged, suggesting that $\mathcal{Z}_t$ alone does not satisfy effectiveness. 

Manipulating only on $\mathcal{C}$ cannot successfully edit all semantics either. In Fig.~\ref{figure3}, we illustrate a failure case where we aim to modify the semantic `cover' in a source image originally labeled `no coat'. By only manipulating $z_t$, the `cover' semantic cannot be correctly edited to `purple coat'. However, when we reweigh the $z_t$s of the source image along with those of images containing a purple coat, we successfully achieve the editing target.

\textbf{Feasibility of linearly manipulating semantics in the latent space.} The disentanglement properties of the diffusion transformer not only enable precise semantic editing, but also enable linearly adjusting the editing degree $\alpha$ that a semantic moves toward a specific value. For a given semantic \( s \) and the corresponding editing direction $n\in \mathbb{R}^{md}$, which defines a hyperplane boundary \( H \subset \mathbb{R}^{md} \) given by \( H = \{ h \in \mathbb{R}^{md} : h \cdot n = 0 \} \), linear manipulation could be performed near $H$, and achieve fine-grained control~\cite{shen2020interpreting}. 

Specifically, in text-to-image diffusion models, text prompts are encoded to token-type embeddings as illustrated in Sec.~\ref{Preliminary}, where each semantic is encoded to different subspaces $C_i\in \mathbb{R}^d,i\in[1,...,m]$.\footnote{We assume each text token corresponds to a specific semantic, and subspaces in \( c_1, c_2, \dots \) correspond to semantics in \( S_1, S_2, \dots \) in a 1-to-1 mapping. Without loss of generality, we disregard the possibility that the position of each semantic's text token may vary across different prompts.} Then we can obtain editing directions $n_i$ of them for manipulations of the text representation on these subspaces. The feasibility of manipulations near the boundary hyper-plane is guaranteed by the following proposition:

\begin{proposition}
\label{threshold}
Let the editing directions $\mathbf{n}_1, \mathbf{n}_2, \dots, \mathbf{n}_m \in \mathbb{R}^d$ be unit vectors, i.e., $\|\mathbf{n}_i\| = 1$ for all $i = 1, \dots, m$. Define $\mathbf{n}_i^{\text{ext}} \in \mathbb{R}^{md}$ as the extension of $\mathbf{n}_i$ into $\mathbb{R}^{md}$ by placing $\mathbf{n}_i$ in the $i$-th block of $d$ coordinates and filling the rest with zeros. Let $\mathbf{z} \sim \mathcal{N}(0, \mathbf{I}_{md})$ be a multivariate random variable. Then we have:
\[
\mathbb{P}\left( \left|\sum_{i=1}^{m} \mathbf{n}_i^T \mathbf{z}_i \right| \leq 2\alpha \sqrt{\frac{d}{d - 2}} \right) \geq \left( \left( 1 - 3e^{-cd} \right) \left( 1 - \frac{2}{\alpha} e^{-\alpha^2 / 2} \right) \right)^m,
\]
for any $\alpha \geq 1$ and $d \geq 4$. Here, $\mathbb{P}(\cdot)$ stands for probability and $c$ is a fixed positive constant.
\end{proposition}

This proposition establishes a bound on how close the vector should be to the hyper-plane. It's important to note that the editing degree should not be excessively large, as increasing it too much, as discussed in Sec.~\ref{boundary and bidirection}, can lead to incorrect image generation or even cause the model to collapse.

\subsection{Probing Analysis on Disentanglement Mechanism}
In this section, we will demonstrate the potential origin of the disentanglement properties of diffusion transformer's latent space, as well as why the representations of classical UNet-based diffusion models are always entangled.

\begin{figure}
    \centering
    \includegraphics[width=1\linewidth]{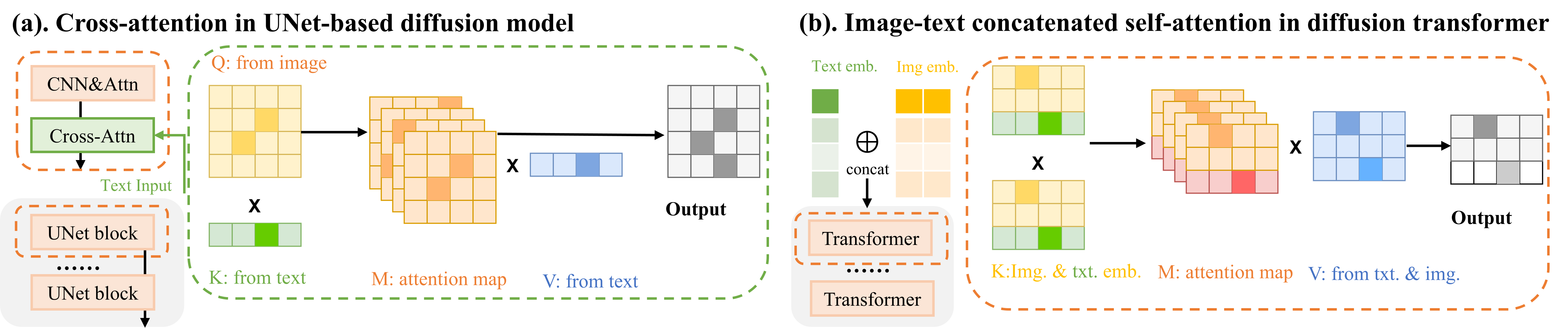}
    \caption{(a) Image-text cross-attention mechanism in UNet-based diffusion models. (b) Self-attention mechanism in the diffusion transformer, where image and text embeddings are concatenated and jointly flowed into the attention block.}
    \label{fig:attention}
\end{figure}

We start from different text conditioning approaches of these two types of models. As shown in Fig~\ref{fig:attention} (a), the denoising block of the diffusion transformer utilizes a self-attention mechanism~\cite{esser2024scaling} that first concatenates image and text embeddings, allowing them to be jointly flowed into self-attention blocks to predict the noise. In contrast, as demonstrated in Fig~\ref{fig:attention} (b), the UNet architecture employs cross-attention layers to introduce text conditions into the denoising process, which might learn entangled mapping from input conditions to image semantics~\cite{liu2024towards}. For example, when generating images with UNet-based diffusion models using the caption \textit{"a $<$color$>$ $<$object$>$"}, the attention map w.r.t. semantic "$<$color$>$" might contain category information of semantic "$<$object$>$". Therefore, when we aim to edit semantic "$<$color$>$", the semantic "$<$object$>$" would be incorrectly modified. 

Building on this insight, to understand the origin of the disentanglement effect, we conduct a probing analysis~\cite{clark2019does,liu2019linguistic} to explore the differences between the self-attention and cross-attention mechanisms in guiding image generation, as utilized in transformer-based and UNet-based diffusion models, respectively. 
Specifically, we generate images using the caption "a $<$color$>$ $<$object$>$" and analyze the attention maps corresponding to each token in the intermediate layers. We then train classifiers using the attention maps of "$<$color$>$" and apply these classifiers to the attention maps of "$<$object$>$. If classifiers successfully identify color information within the "$<$object$>$" attention maps, the attention map may contain category information related to "$<$color$>$". 

\begin{figure}
    \centering
    \includegraphics[width=1\linewidth]{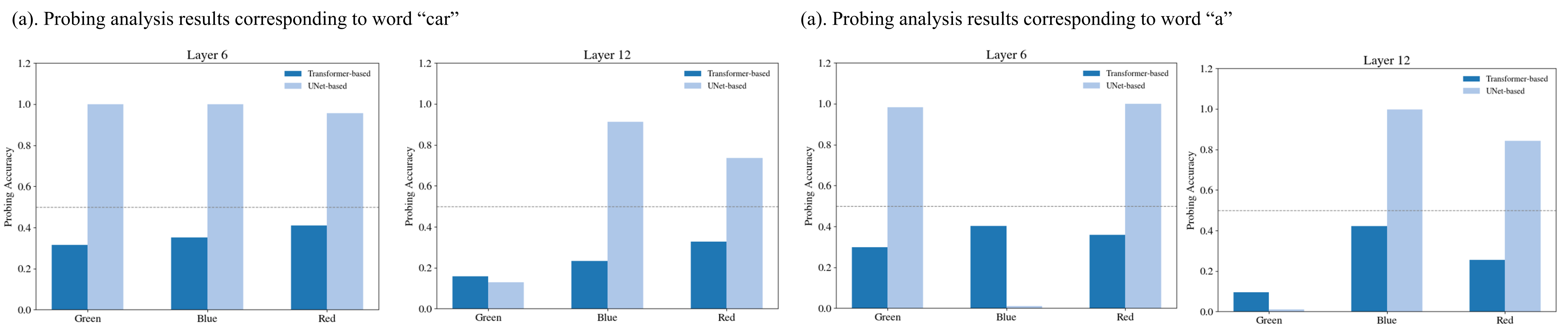}
    \caption{Results of the probing analysis. We provide ratios of attention maps from different models being classified into various color categories. when tested against corresponding color classifiers.}
    \label{fig:linear prob}
\end{figure}

\textbf{The attention map for a specific semantic in the diffusion transformer does not contain category information from other semantics.} As illustrated in Fig.~\ref{fig:linear prob}, the ratio of the diffusion transformer's attention map for ``$<$car$>$'' being classified with a specific color is near 0.5, indicating that it does not effectively encode color category information. This contrasts with the attention map of ``$<$car$>$'' in UNet-based diffusion models, which shows higher tendency in identifying color categories. This observation suggests that the disentanglement property in diffusion transformers stems from their image-text joint self-attention mechanism, leading to less entanglement in the representations corresponding to each semantic. Further experimental details and results are provided in Appendix~\ref{appendix analysis study}.

\section{Method}
In this section, we will first introduce our \textbf{S}emantic \textbf{D}isentanglement m\textbf{E}tric (SDE) metric for measuring the disentanglement degree. Then we will propose our \textbf{E}xtract-\textbf{M}anipulate-\textbf{S}ample (EMS) method for precise and fine-grained image editing.

\subsection{Semantic Disentanglement Metric}
\label{SDE}
Based on disentanglement properties demonstrated in Sec.~\ref{exploration}, we propose a metric that can be automatically computed to measure the degree of disentanglement of the latent space of text-to-image diffusion models.

As we mentioned in Sec.~\ref{disentangle space sec}, a disentangled semantic representation space $S'$ should satisfy decomposability and effectiveness properties, where we can locally manipulate certain subspaces to effectively modify specific semantics without altering other semantics. Building on this insight, we propose to evaluate the disentanglement degree of a semantic representation space using an image-editing-based approach. Specifically,  we aim to examine the Property~\ref{remark decompose} and Property~\ref{remark effectiveness} proposed in Sec.~\ref{disentangle space sec}: (1). \textbf{Effectiveness}: Can we efficiently edit certain semantics on the latent space? (2). \textbf{Decomposable}: Can we avoid incorrect modifications on semantics irrelevant to the editing task? 

Consider an image $x$ with a binary semantic $s\in [0,1]$ that will be lost in the forward process, which is labeled as $y_0=0$. We encode $x$ to latent space to get image embedding $z$, and forward $z$ to obtain $z_t$. We then denoise $z_t$ in the reverse process, and decode it to get the reconstructed image. To ensure effective editing, the reconstructed image \( \tilde{x} \), conditioned on \( y_1 = 1 \), should differ significantly from \( \hat{x} \), which is recovered under the condition \( y_0 = 0 \). This difference is measured by comparing their respective distances to the original \( x \). Simultaneously, to maintain the decomposability property, the distance between \( x \) and \( \tilde{x} \) should be small, ensuring that task-irrelevant semantics remain unchanged during the editing process.
\begin{remark}
    Given a diffusion model with a encode $f$ which turns a given image into the noised image embedding, and a decode function $h$ which turns noised image embedding to the recovered image, an image labeled with $y\in[0,1]$ of semantics $s$, and text embeddings $c, \tilde{c}$ encoded from text prompts correspond to $y$ and $1-y$. The \textbf{S}emantic \textbf{D}isentanglement m\textbf{E}tric (SDE) could be written as: 
    \begin{equation}
    \footnotesize
    \text{SDE}=\frac{||x-h(f(x,t), c, t)||_2}{||x-h(f(x,t), \tilde{c},t)||_2}+||x-h(f(x,t), \tilde{c}, t)||_2
    \end{equation}
    where $t$ is the forward time-step.
\end{remark}


\subsection{Extract-Manipulate-Sample Framework}
\label{sec ems}
\begin{figure*}[htp] 
    \centering
    \includegraphics[width=1\textwidth]{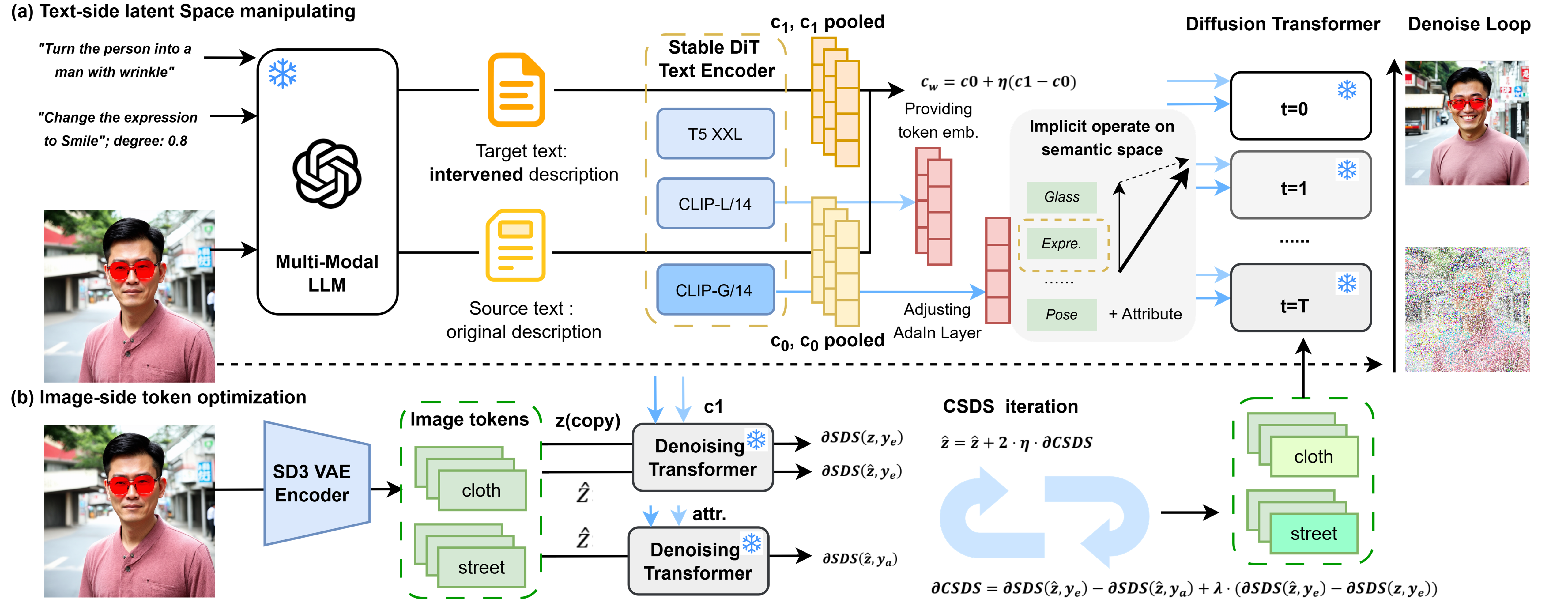} 
    \caption{(a). \textbf{Extract}: Given a source image, we first utilize a multi-modal Large Language Model (LLM) to get a source prompt that describes semantics of the image, and a target prompt that describes the desired edited images. (b). \textbf{Manipulate}: On the text side, we manipulate text embeddings with the editing direction, which can be obtained by subtracting the embeddings encoded from the source prompt and target prompt. (c). \textbf{Sample}: On the image side, we leverage a constrained score-distillation-sampling method to iteratively update the image latents with Eq.~\ref{csds}.
    }
    \label{fig:method}
\end{figure*}

In this section, we propose our \textbf{E}xtract-\textbf{M}anipulate-\textbf{S}ample (EMS) framework that leverage semantic disentanglement properties for precise and fine-grained image editing. The whole framework can be seen in Fig.~\ref{fig:method}.

\textbf{Extract.} As shown in Sec.~\ref{disentangle space sec}, semantics that are lost in the forward process would be randomly reconstructed during the reverse process. To help condition the recovery of semantics those are lost in the forward process, given an image $x$, we need to first obtain a text description $t_0$ of it. In practice, we utilize a multi-modal LLM~\cite{achiam2023gpt} to extract the source text prompt $t_0$. Similarly, we use a multi-modal LLM to generate the text prompt $t_1$ that describes the desired image. We instruct the LLM to only replace the word related to the editing target, while keeping the other words in the text description unchanged. We then encode them with text encoders of the diffusion transformer to get text embedding and its pooled one $c_0, c_{0,pool}$ and $c_1,c_{1,pool}$ respectively.

\textbf{Manipulate.} Motivated by Sec.~\ref{disentangle space sec}, the latent space of text embeddings is highly decomposable, and the editing directions are explicit. Therefore, for the given semantic to be edited, we first try to modify the semantic via employing the text-side manipulation. As illustrated in Fig.~\ref{figure2}, given the ratio $\alpha$ for modifying the semantic $s$, we construct the editing direction $n$ here with $n=c_1 - c_0$, and manipulate text embeddings with: $\tilde{c}=c+\alpha \cdot n, c=c_0$. Similarly, we manipulate the pooled embedding with $\tilde{c_{pool}}=c_{pool}+\alpha \cdot (c_{1,pool} - c_{0,pool}), c_{pool}=c_{0, pool}$.

\textbf{Sample.} As shown in Sec.~\ref{disentangle space sec}, manipulation on text embedding space can not always successfully edit the target semantic. While increasing the noise level in the forward process and leveraging the phenomenon of semantic loss can enable editing of more semantics via text manipulation, it may also result in the randomly recovery of detailed semantics that are irrelevant to the editing task and are not identified in the extract stage. This brings a trade-off on precise editing and efficient editing. To solve this dilemma, during a editing task, we would first try image-side manipulation, when text-based techniques fail, indicating that the target semantic may not have been lost in the forward process, we need to manipulate the noised image embeddings. However, for the noised image embeddings $z_t$, assuming there is no reference image or embedding, we lack an explicit editing direction as in the example case in Fig.~\ref{figure3}. To address this problem, we employ score-distillation sampling to manipulate the image embeddings. Specifically, we follow~\cite{hertz2023delta,poole2022dreamfusion} to iteratively sample the image embedding $z_t$ to get $z_t'$.

During score-distillation sampling, we observed that inputting only the text embedding $c_{\text{attr}}$ corresponding to the desired semantic value (e.g., "purple coat") in text manipulation, can reduce the difficulty in modifying that semantic. This insight leads us to minimize the distance between the predicted noise $\epsilon_{\tilde{c}}(z_t')$, conditioned on $\tilde{c}$, and $\epsilon_{\text{attr}}(z_t')$, conditioned on $c_{\text{attr}}$. However, minimizing this distance might affect semantics irrelevant to the semantic to be edited as we illustrate in the analysis experiment. Hence, we add a constrained term which aims to reduce the distance between $\epsilon_{\tilde{c}}(z_t')$ and noise $\epsilon_{\tilde{c}}(z_t)$ predicted using $z_t$ and $\tilde{c}$ as input. Finally, the image gradient of the constrained score-distillation sampling could be:
\begin{equation}
\footnotesize
\label{csds}
    \partial CSDS=\partial (||\epsilon_{\tilde{c}}(z_t')-\epsilon_{\text{attr}}(z_t')||^2_2+\lambda\cdot ||\epsilon_{\tilde{c}}(z_t')-\epsilon_{\tilde{c}}(z_t)||^2_2)
\end{equation}
, where $\lambda$ is the hyper-parameter that adjusts the degree to which the manipulated image embedding $z_t'$ should be close to the original $z_t$.

In score-distillation-sampling, we do not require the gradient of the diffusion transformer. Hence we can directly compute $\partial CSDS$, and update the image embedding with: $z_t'=z_t'- \eta \cdot \partial CSDS$, where $\eta$ refers to the step size for the update loop.

\section{Experiment}
\subsection{Experiment Setting}
\textbf{Implementation Detail}. Our experiments are manly conducted on Stable Diffusion V3~\cite{esser2024scaling}, the most popular open-source transformer-based Text-to-Image (T2I) diffusion model. 
We follow the implementation of Stable Diffusion V3 of the Huggingface Diffusers library~\cite{von-platen-etal-2022-diffusers}, and use the provided checkpoint with the model card "stabilityai/stable-diffusion-3-medium-diffusers". For qualitative results, we set the classifier-free-guidence (CFG) scale to 7.0. For the image editing task, we set the CFG scale to 7.5 and the total sampling steps to 50, we first forward the image to $75\%$ of the total timesteps, and then conduct the reverse process. For our UNet-based T2I model, we utilize Stable Diffusion V2.1 from Huggingface Diffusers~\cite{von-platen-etal-2022-diffusers}, which is the most widely used UNet-based T2I diffusion model. We set the hyperparameters the same as those mentioned above. During our experiments, we kept diffusion models frozen for zero-shot image-editing task. For the implementation of our method, we utilize the GPT-4~\cite{achiam2023gpt} to extract text description of given images. Details of the prompts input to the multi-modal LLM to obtain the text descriptions can be found in the appendix.

\textbf{Baseline.} Since text-guided zero-shot fine-grained image editing is a novel task, we adapted several existing image editing methods with text-to-image diffusion models. Specifically, we utilize DiffEdit~\cite{couairon2022diffedit}, Pix2Pix~\cite{brooks2023instructpix2pix}, and MasaCtrl~\cite{cao2023masactrl} as our baselines. Most of these methods rely on UNet-specific network structures or embeddings, so we implemented their backbones using a UNet-based diffusion model. To enable fine-grained image editing, we applied similar text manipulation techniques to the text embeddings that guide the reverse process of image editing.

\textbf{Qualitative Evaluation.} We conducsed a comprehensive qualitative assessment of EMS using a diverse array of real images spanning multiple domains. Our evaluation employed simple text prompts to describe various editing categories, including but not limited to style, appearance, shape, texture, color, action, lighting, and quantity. These edits were applied to a wide range of objects, such as humans, animals, landscapes, vehicles, food, art, and moving objects elements. To demonstrate the necessity of image-side manipulation, we compared EMS's performance with text-only embedding space manipulations. Table \ref{tab:table2} presents a checklist of the qualitatively investigated image editing tasks along with illustrative examples. For our base images, we generated high-resolution samples using Stable Diffusion 3. We then applied EMS with 3 different editing degrees to showcase fine-grained editing capabilities and the controllability of the editing degree. 

Additionally, we compared the fine-grained editing results of our EMS method with those of the baseline models. We adapted the baselines to the same experimental pipeline to conduct fine-grained image editing. Like EMS, these methods were applied with 3 different editing degrees to demonstrate their capabilities. The comparison highlights EMS's superior ability to achieve precise, fine-grained edits and control the degree of modification.

\textbf{Benchmark Construction and Quantitative Evaluation}
Despite recent advancements in diffusion-based image editing, there has been a notable absence of well-labeled, objective, and fine-grained evaluation benchmarks, particularly for zero-shot editing. While \cite{kawar2023imagic} introduced TEDBench, a 100-image benchmark for fine-grained image editing, its reliance on human evaluation makes it time-consuming and costly to replicate across different models. Similarly, \cite{ju2023direct} developed PIEBench with both human and machine evaluations, but it lacks comprehensive assessment of fine-grained editing, especially regarding the degree of edit features.
To address these limitations and quantify the effectiveness of our EMS framework, we introduce ZOFIE (Zero-shot Open-source Fine-grained Image Editing benchmark). ZOFIE is the first diffusion-based image editing benchmark that incorporates both human subjective and automatic objective machine evaluations, assessing image quality, controllability of fine-grained edits, image-text consistency, background preservation, and semantic disentanglement.
ZOFIE comprises 576 images evenly distributed across 8 editing types and 8 object categories (humans, animals, landscapes, vehicles, food, art, moving objects, and sci-fi elements). Each benchmark element includes source prompt, edit prompt, editing feature, object class, object region of interest mask, and background mask annotations.
For human subjective evaluation, we conducted an Amazon Mechanical Turk user study following the standard Two-Alternative Forced Choice (2AFC) protocol. We gradually increased the editing strength of EMS and other baseline methods, producing three consecutive image edits. Four participants with diverse backgrounds were presented with our edits and baseline edits, and asked to select the superior set of images based on editing quality, background preservation, and edit-prompt-image consistency. In total, 20,736 results were collected.
For automatic and objective evaluation, we employed six metrics across four aspects:
\begin{itemize}
    
\item Background preservation: PSNR, LPIPS, and SSIM
\item Text-image consistency: CLIPScore
\item Fine-grained editing effectiveness: MLLM-VQA score (Multimodal LLM-based Visual Question Answering score) This metric uniquely leverages advanced multimodal language models to assess the effectiveness of gradual fine-grained editing. By posing targeted questions about consecutive edits to a large language model, we quantify the perceptibility and smoothness of incremental changes in the edited images.
\end{itemize}
The CLIP score was calculated between the edit prompt and the edited images. For the MLLM-VQA score, we input pairs of images with consecutive editing strengths into GPT-4 and posed the question: "Answer only in yes or no, does the second image reflect a gradual change of [edit feature] compared to the first image?" We ran this process five times per sample and calculated the percentage of "yes" responses across all runs and samples.

\begin{figure*}[t!] 
    \centering
    
    \includegraphics[width=1\textwidth]{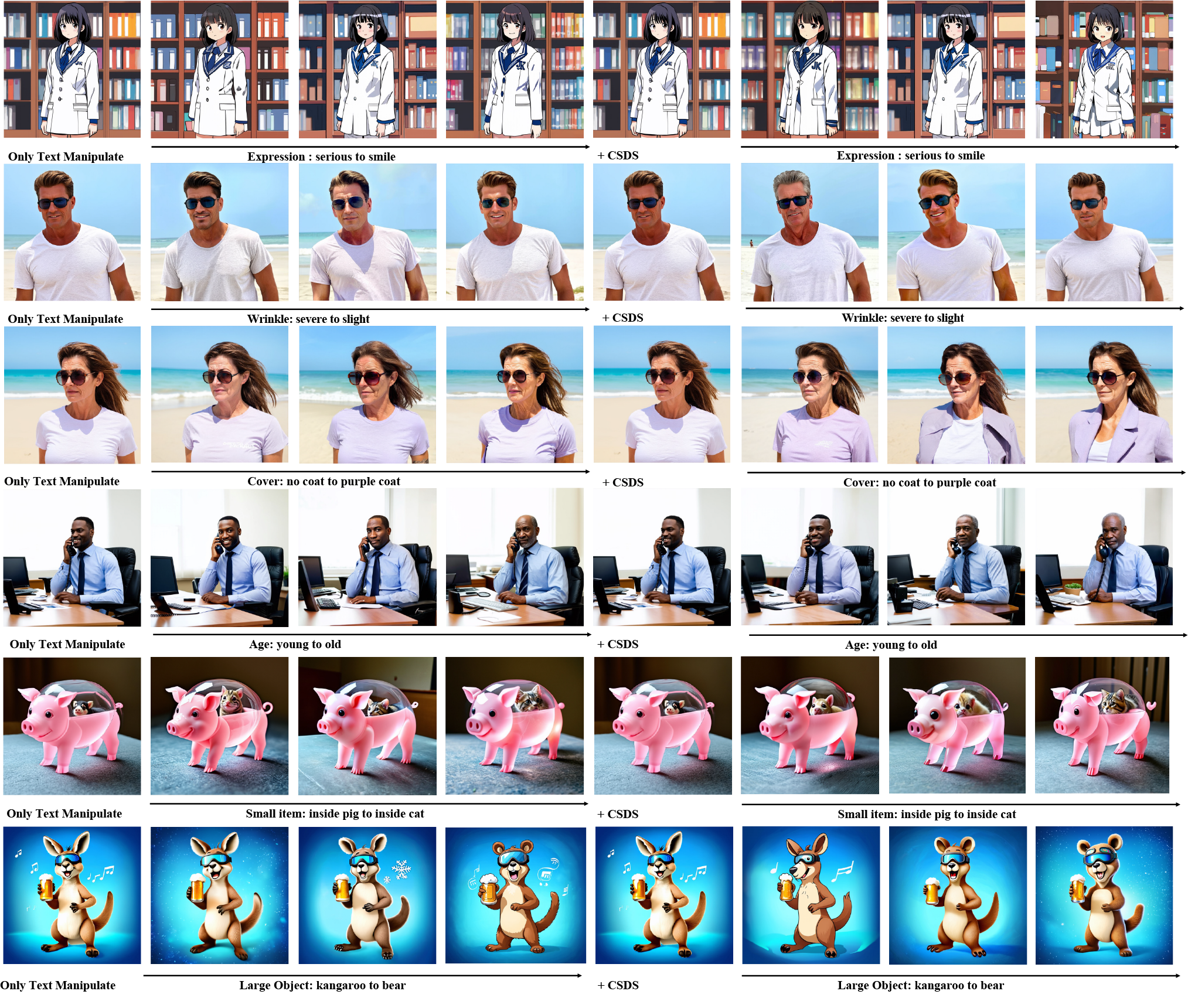} 
    \caption{Fine-grained editing results of person-based and object-based semantics. We utilize diffusion transformers to gradually modify the considered attributes, with our EMS method and text-manipulation-only method. Both text-side manipulation and our proposed EMS method can handle these editing cases.
    }
\label{fig2}
\end{figure*}

\textbf{Evaluation on semantic disentanglement degree.} We employ the SDE metric proposed in Sec.~\ref{SDE} to quantitatively assess the disentanglement degree of text-to-image diffusion models. Specifically, we calculate the SDE for both the UNet-based diffusion model and the diffusion transformer using the CeleBA~\cite{liu2015deep} dataset. CeleBA is a widely used benchmark for evaluating the generative capabilities of diffusion models~\cite{rombach2022high,zhang2024distributionally}, associates each image with an attribute vector, making it particularly suitable for measuring the disentanglement degree with the SDE metric. We randonly sampled 2000 samples from the whole dataset, and test the disentanglement degree of the models based on age, gender, expression, hair, eyeglasses, and hat attributes.

\begin{table}[h] 
\centering 

\caption{Checklist of the qualitatively investigated editing task. Text tokens are more efficiently in controling semantics related to person and object than attribute and texture.}
\scalebox{0.71}{
\begin{tabular}{@{} l lll lll lll lll}
\toprule
\textbf{method} &\multicolumn{3}{c}{\textbf{Attribute}} &\multicolumn{3}{c}{\textbf{Texture}} &\multicolumn{3}{c}{textbf{Person}}  &\multicolumn{3}{c}{\textbf{Object}}\\
\cmidrule(l){2-4} \cmidrule(l){5-7} \cmidrule(l){8-10} \cmidrule(l){11-13} 
& Category & Example & &Category & Example & &Category & Example & &Category & Example & \\

\midrule
EMS & quantity& tiny$\to$huge& $\checkmark$& light & bright$\to$dark & $\checkmark$& smile & slight$\to$intense & $\checkmark$ & cover & none$\to$coat & $\checkmark$\\
EMS & shape & rect.$\to$spher.&$\checkmark$ & surface& uneven$\to$iron & $\checkmark$ & age & young$\to$old & $\checkmark$ & small item & pig$\to$cat &$\checkmark$\\
EMS & color & blue $\to$ green& $\checkmark$ & style & pencil$\to$painting & $\checkmark$ & wrinkle & severe$\to$slight & $\checkmark$ & large object & kangaroo$\to$bear & $\checkmark$\\

Text-only &quantity& tiny$\to$huge&  & light & bright$\to$dark &  & smile & slight$\to$intense & $\checkmark$ & cover & none$\to$coat &\\
Text-only & shape & rect.$\to$spher.& & surface& uneven$\to$iron & $\checkmark$& age & young$\to$old & $\checkmark$& small item & pig$\to$cat &$\checkmark$\\
Text-only& color & blue $\to$ green& $\checkmark$ & style&pencil$\to$painting & & wrinkle & severe$\to$slight & $\checkmark$& large object & kangaroo$\to$bear & $\checkmark$\\
\bottomrule
\end{tabular}
}
\label{tab:table2}
\end{table}

\subsection{Qualitative Results}

\begin{figure*}[h!] 
    \centering
    
    \includegraphics[width=1\textwidth]{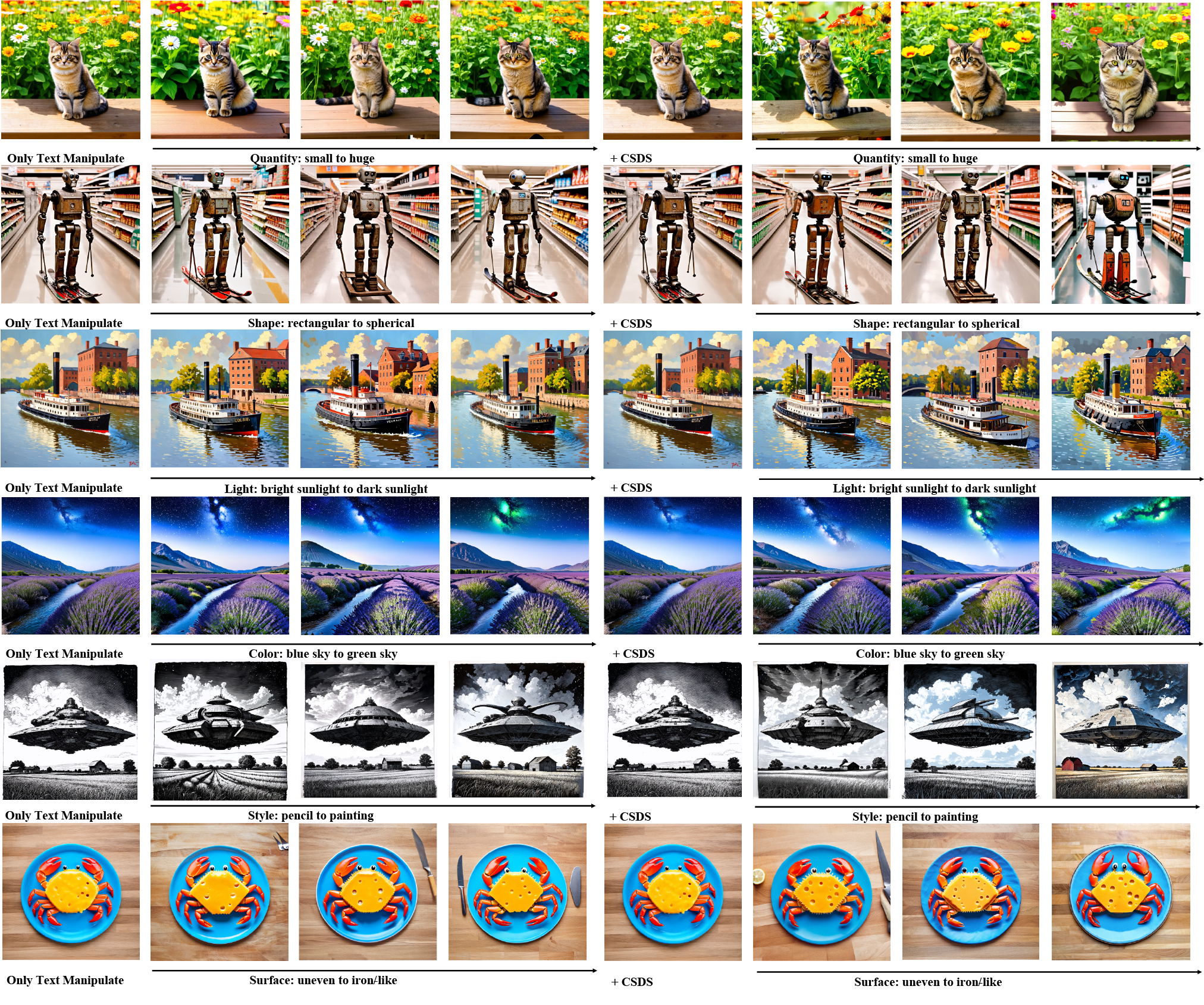} 
    \caption{ Fine-grained editing results of attribute-based and texture-based semantics. We utilize diffusion transformers to gradually modify the considered attributes, with our EMS method and text-manipulation-only method. While text-only method failed to achieve certain editing target in these two types of semantics, our EMS method has successfully addressed this problem.
    }
\label{fig1}
\end{figure*}

\textbf{EMS achieves greater effectiveness in editing target semantics by operating on the entire semantic representation space, whereas text-only manipulation is limited to certain semantics. }Fig.\ref{fig1} and Fig.\ref{fig2} illustrate EMS's ability to apply various editing categories to general input images and texts effectively. Notably, while text-side only manipulation failed to achieve certain tasks, such as "enlarging a cat sitting in the garden" or "adding a purple coat to an American woman by the sea," EMS successfully accomplished these edits in a zero-shot manner by manipulation in the image latent space, without requiring reference images. Furthermore, EMS exhibited smooth semantic editing properties as the editing strength increased. This is evidenced by the gradual enhancement of the target edit from left to right in Fig.\ref{fig1} and Fig.\ref{fig2}, demonstrating the fine-grained control over the editing process.

\begin{figure*}[h!] 
    \centering
    
    \includegraphics[width=1\textwidth]{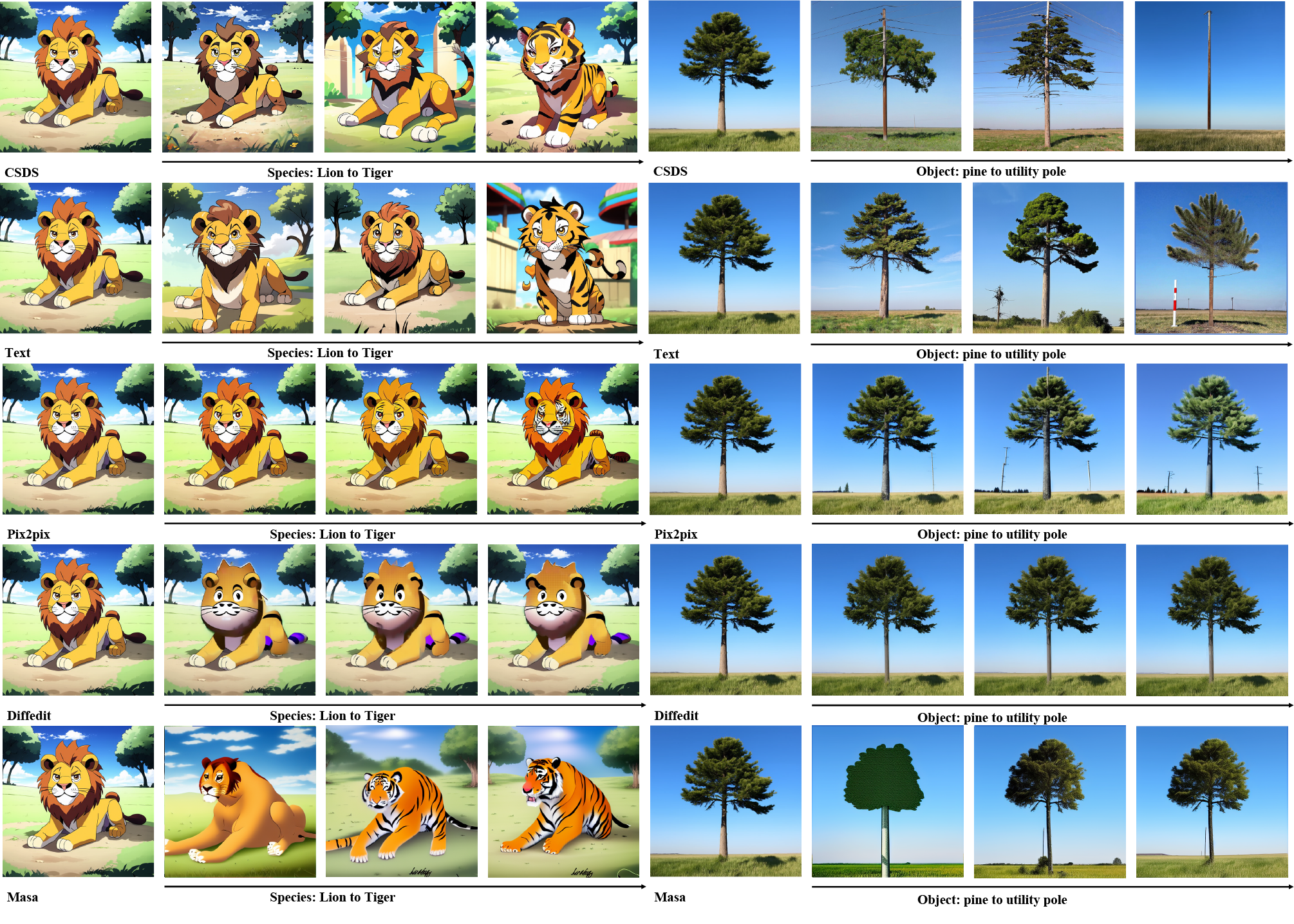} 
    \caption{Comparisons of fine-grained editing results of our EMS method and baseline methods. Our EMS method has successfully achieved progressive transformations from lion to tiger and pine tree to utility pole. While baseline methods fail to achieve the target transformations, often degrading the original image quality significantly or producing unintended effects.
    }
\label{baseline}
\end{figure*}

\textbf{EMS outperforms baseline approaches in achieving more precise and fine-grained image editing.} As illustrated in Fig.~\ref{baseline}, the baseline methods frequently introduce unintended changes to irrelevant semantics, disrupting the overall image coherence. In contrast, EMS excels in making fine-grained adjustments to target semantics, achieving a level of detail and precision that baseline methods struggle to match. This highlights EMS's advanced capability in preserving non-target attributes while effectively executing the intended edits.

\subsection{Quantitative Results}
\textbf{Our EMS method achieve better precise and fine-grained image editing results.} As shown in Table~\ref{quantitative table}, EMS demonstrates a significant advantage in background preservation during edits, with a strong PSNR of 20.8 and the lowest LPIPS score of 133.6. This showcases EMS's ability to decompose and isolate specific semantics, ensuring that areas irrelevant to the editing target remain unaffected by the modifications. Moreover, the high CLIP Similarity Score of 34.3 achieved by EMS highlights its effectiveness in accurately editing target semantics. Finally, the significant advantage observed in the MLLM-VQA metric, where EMS scores 43.8, demonstrates its strength in fine-grained editing, reflecting the model's ability to produce nuanced and precise modifications.

\begin{table}[ht]

    \centering
    \caption{Quantitative Results}
    \resizebox{\textwidth}{!}{%
    \begin{tabular}{lcccccc}
        \toprule
        & \multicolumn{3}{c}{Background Preservation} & \multicolumn{1}{c}{CLIP Similarity Score} &\multicolumn{1}{c}{MLLM-VQA} \\
        \cmidrule(lr){2-4} \cmidrule(lr){5-5} \cmidrule(lr){6-6}
        Method & PSNR $\uparrow$ & LPIPS $\downarrow \times 10^3$ & SSIM $\uparrow \times 10^2$ & Whole $\uparrow \times 10^2$ & Whole $\uparrow \times 10^2$  \\
        \midrule
        P2P~\cite{brooks2023instructpix2pix} & \textbf{21.0} & \underline{139} & \underline{79.5} & 30.1 & 16.7 \\
        DiffEdit~\cite{couairon2022diffedit} & 19.1 & 205.4 & 69.9 & 30.7 & 14.6 \\
        MasaCtrl~\cite{cao2023masactrl} & 16.4 & 195.8 & 68.7 & 28.5 & 18.5 \\
        Prompt Only & 20.2 & 183.5 & 73.9 & \underline{31.2} & \underline{29.2} \\
        Ours & \underline{20.8} & \textbf{133.6} & \textbf{79.6} & \textbf{34.3} & \textbf{43.8} \\
        \bottomrule
    \end{tabular}%
    }
    \label{quantitative table}
\end{table}

\textbf{The diffusion transformer has better semantic disentanglement degree.} As indicated in Table~\ref{sde result}, the diffusion transformer consistently achieves lower SDE scores across various attributes such as age, gender, expression, hair, eyeglasses, and hat, compared to the UNet-based diffusion model. The average SDE score of the diffusion transformer is significantly lower (1.130) compared to that of the UNet-based model (1.302). This suggests that the diffusion transformer has a more disentangled representation space, leading to more precise control during image editing.

\begin{table}[h]
\centering
\scalebox{0.65}{
\begin{tabular}{l lllllll}
\toprule
Model & SDE on Age$\downarrow$ & SDE on Gender$\downarrow$ & SDE on Expression$\downarrow$ & SDE on Hair$\downarrow$ & SDE on Eyeglasses$\downarrow$ & SDE on Hat$\downarrow$ & Avg.\\
\midrule
UNet-based~\cite{rombach2022high} & 1.424 & 1.117 & 1.425 & 1.287 & 1.283 & 1.273 & 1.302\\
Transformer-based~\cite{esser2024scaling} &  \textbf{1.138} & \textbf{1.080} & \textbf{1.151} & \textbf{1.192} & \textbf{1.123} & \textbf{1.096} & \textbf{1.130}\\
\bottomrule
\end{tabular}
}
\caption{The semantic disentanglement metric of the diffusion transformer and the UNet-based diffusion model.}
\label{sde result}
\end{table}

\section{Analysis Study}

\subsection{Extension to Multi-Attribute Editing}
Our EMS method can be extended to multi-attribute editing scenarios. Formally, consider a semantic $s\in \mathcal{S}$ that consists of modular semantic $s_i \in \mathcal{S}_i, i\in [1,...,m]$. The editing direction $n_i$ for each semantic $s_i$ can be obtained by encoding a text prompt that corresponds to the target value of $s_i$. We extend it to the entire space to get a $\mathbf{n}_i^{\text{ext}}\in \mathbb{R}^{md}$. The extended editing directions of different semantics are orthogonal to each other, enabling disentangled editing and facilitating adaptation to multi-semantic editing tasks:

\begin{proposition}
    Let the editing directions $\mathbf{n}_1, \mathbf{n}_2, \dots, \mathbf{n}_m \in \mathcal{C}_1,...,\mathcal{C}_m$ of semantics be unit vectors, i.e., $\|\mathbf{n}_i\| = 1$ for all $i = 1, \dots, m$. Define $\mathbf{n}_i^{\text{ext}} \in \mathbb{R}^{md}$ as $\mathbf{n}_i^{\text{ext}} = (0_d, \ldots, 0_d, \mathbf{n}_i, 0_d, \ldots, 0)$, where $\mathbf{n}_i$ occupies the $i$-th block of $d$ coordinates. Then, $\{\mathbf{n}_i^{\text{ext}}\}_{i=1}^{m}$ forms an orthogonal set in $\mathbb{R}^{md}$, i.e., for $i \neq j$, $\mathbf{n}_i^{\text{ext}} \cdot \mathbf{n}_j^{\text{ext}} = 0$.    
\end{proposition}

For multi-attribute editing scenarios, consider editing directions \( \mathbf{n}_1, \mathbf{n}_2, \dots, \mathbf{n}_m \in \mathbb{R}^d \) corresponding to semantics \( s_1, s_2, \dots, s_m \), where the orthogonality of \( \mathbf{n}_i \) and \( \mathbf{n}_j \) (i.e., \( \mathbf{n}_i \perp \mathbf{n}_j \)) is ensured by the decomposability property. For editing degrees \( \alpha_1, \alpha_2, \dots, \alpha_m \), these can be applied individually to \( \mathbf{n}_1, \mathbf{n}_2, \dots, \mathbf{n}_m \), and then extended into a combined editing direction \( \mathbf{n}^{\text{ext}} = (\alpha_1 \cdot \mathbf{n}_1, \alpha_2 \cdot \mathbf{n}_2, \dots, 0_d) \in \mathbb{R}^{md} \), where \( m \) represents the total number of semantics involved.

Specifically, for single-semantic manipulation on the text side, we compute the editing direction $n$ by $c_1-c_0$, where $c_1,c_0$ is the embedding of extracted text description of the source image and desired image. Therefore, we used the following formula for text-side manipulation:
\[
\mathbf{c} = \mathbf{c}_0 + \alpha (\mathbf{c}_1 - \mathbf{c}_0)
\]
where $\mathbf{c}_0$ is the original embedding, $\mathbf{c}_1$ is the target embedding, and $\alpha$ is the editing strength.

For multi-attribute editing, we can extend this to:
\[
\mathbf{C} = \mathbf{C}_0 + \Lambda (\mathbf{C}_1 - \mathbf{C}_0)
\]
Here, $\Lambda$ is a diagonal matrix indicating the editing degree of multiple attributes. 
This formulation leverages the decomposability of the disentangled semantic representation space, allowing us to edit multiple attributes simultaneously without interference.

\begin{figure*}[t!] 
    \centering
    
    \includegraphics[width=1\textwidth]{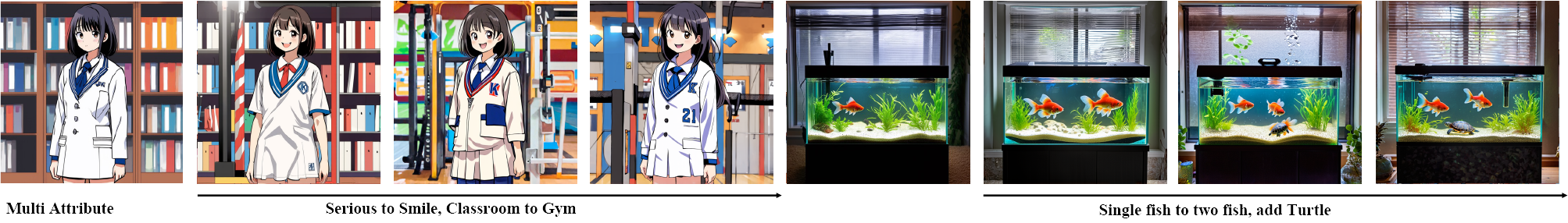} 
    \caption{Extension to multi-attribute fine-grained editing. EMS method has successfully edited multiple attributes simultaneously.
    }
\label{multi-attr fig}
\end{figure*}

As shown in Fig.~\ref{multi-attr fig}, our EMS has successfully achieved fine-grained multi-attribute editing.

\subsection{Bidirectional Editing and Limitations}
\label{boundary and bidirection}
Our method also supports bidirectional editing, which can be theoretically justified by theorems proposed in ~\cite{shen2020interpreting}. When the latent vector moves across the hyperplane formed by the editing direction based on the target value of the semantic (e.g., semantic value "smile" of the semantic "expression", as discussed in Sec.~\ref{disentangle space sec}), the sign of the distance between the latent representation and the hyperplane flips. This allows us to both increase and decrease the intensity of an given semantic value.

Formally, given an normal vector $n_i$ of editing direction for semantic $s_i,i\in[1,...,m],m\in N^+$, and a latent representation $z$, the signed distance to the hyperplane is given by: $D = n_i^T z$. As $z$ moves in the direction of $n_i$ or $-n_i$, the disentance $d$ becomes positive or negative, respectively, allowing bidirectional control of the attribute.

\begin{figure*}[htp] 
    \centering
    
    \includegraphics[width=1\textwidth]{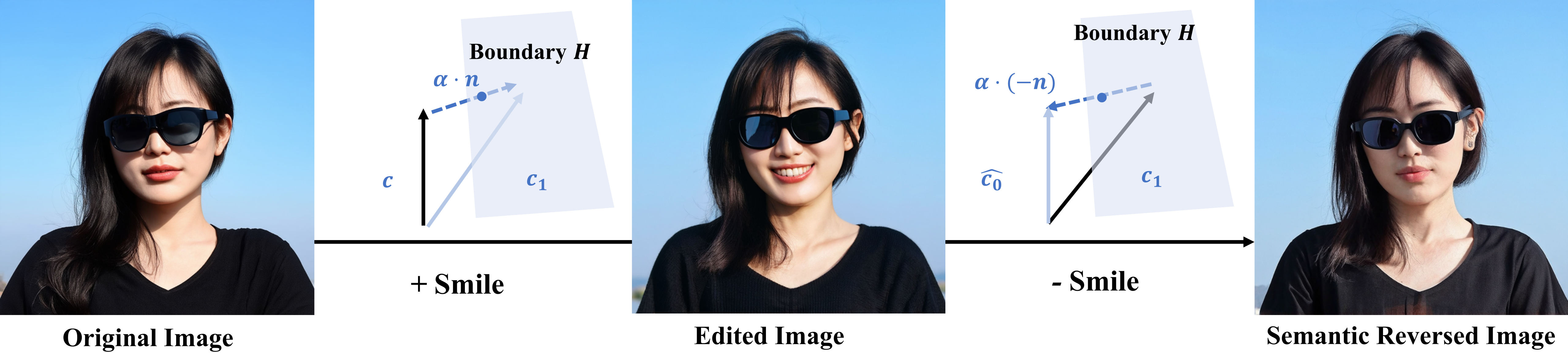} 
    \caption{Edited semantics can be reverted to their original values. The disentanglement property allows precise editing of the target semantic without affecting others, enabling us to use the same direction that was used to edit the semantic to reverse it back to its original value.
    }
\label{ablation_reverse}
\end{figure*}

As shown in Fig.~\ref{ablation_reverse}, we have achieved bi-directional image editing across the boundary defined by the editing direction. We first add "smile" to the expression semantic, and then remove the 

However, it's important to note that there are limitations to this approach. As demonstrated in our experiments (figure place holding), when we move the latent vector $z$ too far from the hyperplane, exceeding a certain threshold, the probability of the projection falling near the hyperplane diminishes rapidly. This is consistent with the concentration of measure phenomenon in high-dimensional spaces \cite{shen2020interpreting}. In our work, the threshold can be given by Proposition~\ref{threshold} in Sec.~\ref{disentangle space sec}



\begin{figure}
    \centering
    \includegraphics[width=\linewidth]{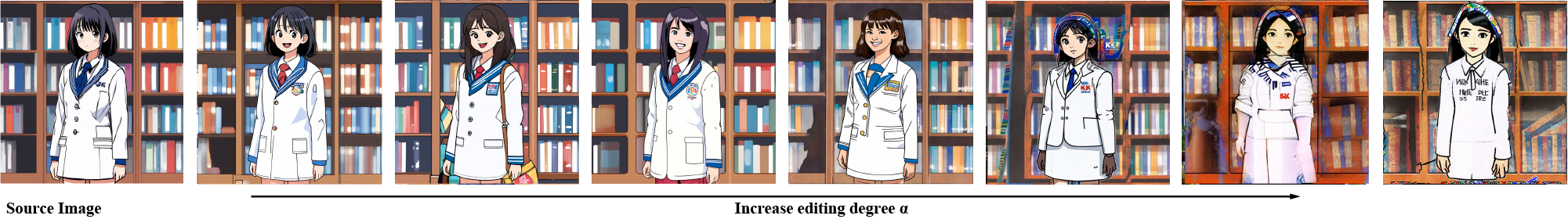}
    \caption{Linear manipulation is effective near the boundary defined by the editing direction. However, if the edit degree exceeds the threshold set by Proposition.(\ref{threshold}), the resulting image may become corrupted.}
    \label{fig:corrupt}
\end{figure}

When the threshold is exceeded, we observe a loss of controlled attribute editing. For example, in our experiments with facial attributes, we found that extreme modifications to the "expression" semantic began to affect unrelated features such as ethnicity as shown in Fig.~\ref{fig:corrupt}.

This phenomenon highlights the importance of careful calibration of the editing strength $\alpha$ (or the diagonal elements of $\Lambda$ in multi-attribute editing) to maintain precise control over the desired attributes while preserving other semantic features of the image.

\subsection{Hyperparameter Analysis}

\begin{figure*}[htp] 
    \centering
    
    \includegraphics[width=1\textwidth]{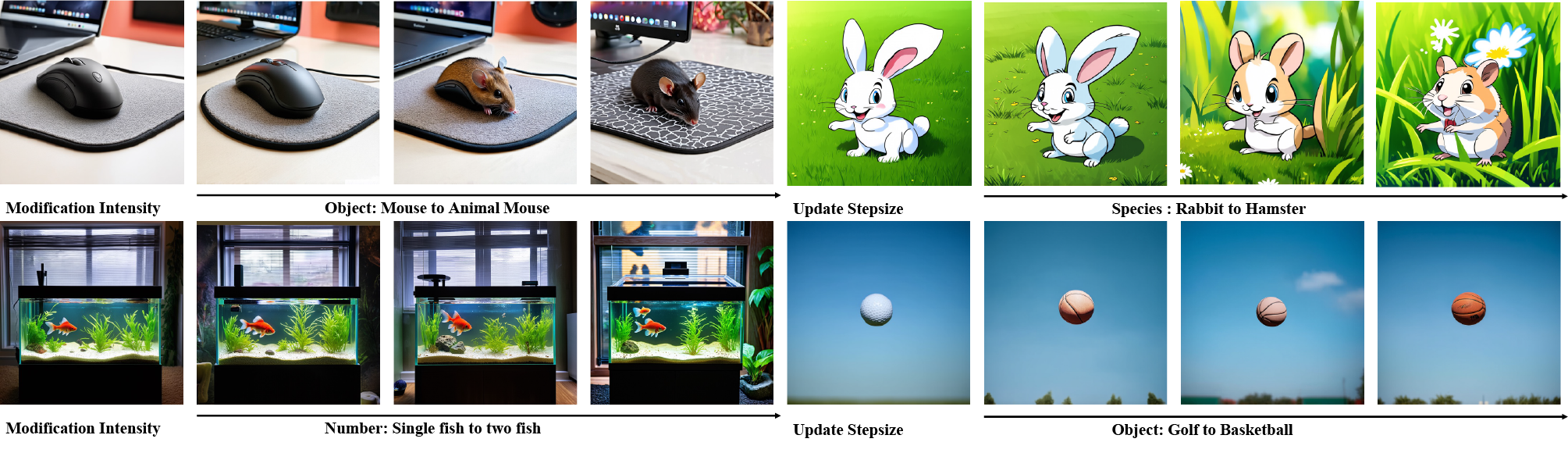} 
    \caption{Ablation study on hyperparameters in our EMS method. The figure demonstrates the effects of modification intensity ($\lambda$) and update stepsize on various editing tasks. These examples highlight how careful tuning of these hyperparameters allows for precise control over edit strength, content preservation, and the balance between desired modifications and maintaining original image semantics.
    }
\label{hyper-param 1}
\end{figure*}

In addition to the editing strength $\alpha$, two crucial hyperparameters in our EMS method are the update stepsize and the modification intensity $\lambda$ used in the constrained score-distillation sampling (CSDS) process.

\textbf{Update Stepsize:} The update stepsize $\eta$determines the magnitude of each update during the Score-Distillation Sampling process. We found that a carefully chosen stepsize is essential for achieving smooth and controlled edits. Too large a stepsize can lead to overshooting and instability in the editing process, while too small a stepsize can result in slow convergence and subtle edits.

In our experiments, we used an adaptive stepsize strategy, starting with a relatively large value of 0.1 and gradually decreasing it to 0.01 as the progresses. This approach allowed for rapid initial progress followed by finer adjustments, resulting in more precise edits.

\textbf{Modification Intensity $\lambda$:} The modification intensity $\lambda$ in our CSDS objective (Eq.~\ref{csds}) plays a crucial role in balancing between achieving the desired edit and preserving the original image content. It controls the trade-off between modifying the target attribute and maintaining other semantic features of the image.

We empirically found that setting $\lambda$ in the range of 0.1 to 1.0 generally produces good results across various editing tasks. Specifically:

\begin{enumerate}
    \item Higher $\lambda$ values (0.7 - 1.0) result in more conservative edits that better preserve the original image structure but may require more iterations to achieve the desired effect.
    \item Lower $\lambda$ values (0.1 - 0.3) allow for more aggressive edits but may occasionally introduce unintended changes in other attributes.
\end{enumerate}

The optimal value of $\lambda$ can vary depending on the specific attribute being edited and the desired strength of the edit. For example, we observed that edits involving more localized changes (such as modifying facial features) generally benefit from higher $\lambda$ values, while global edits (like changing the overall style or lighting) can tolerate lower $\lambda$ values.

Figure~\ref{hyper-param 1} illustrates the effect of different $\lambda$ values on the editing process for a sample image, demonstrating the trade-off between edit strength and content preservation.

These findings not only provide insight into the behavior of our EMS method but also shed light on the structure of the latent space in diffusion transformers. Understanding these properties and limitations is crucial for developing more robust and controllable image editing techniques in future work. Further research could explore automatic adaptation of these hyperparameters based on the specific editing task and image content, potentially leading to more robust and user-friendly image editing systems.

\subsection{Detailed Analysis on Semantic loss}

\begin{figure*}[htp] 
    \centering
    
    \includegraphics[width=1\textwidth]{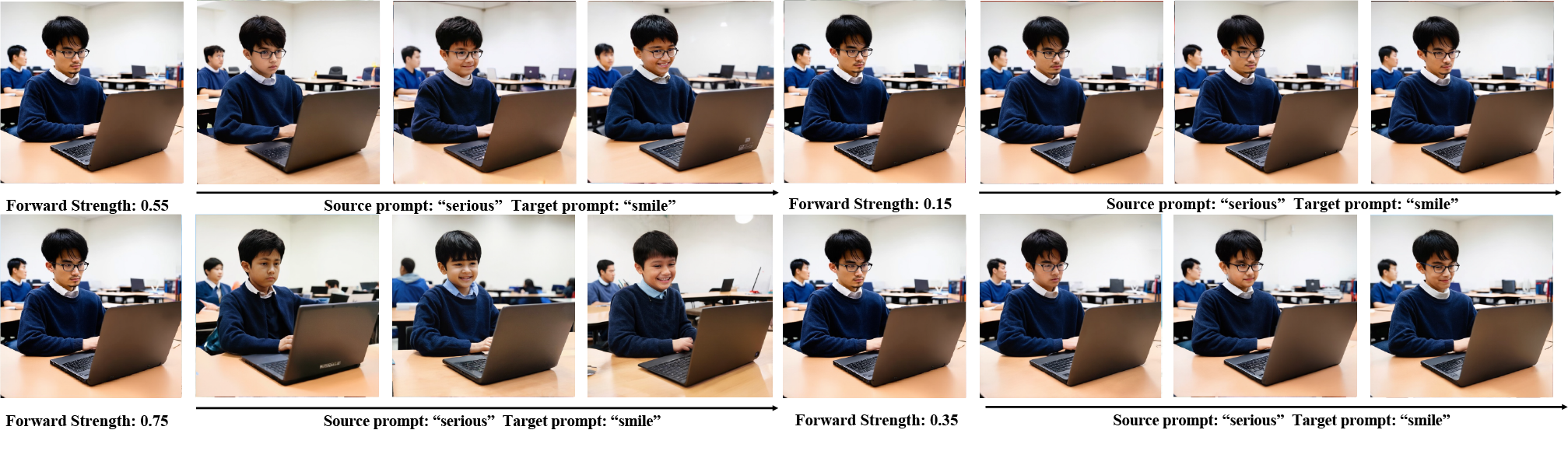} 
    \caption{Semantics are gradually lost during the forward process. If not conditioned, these lost semantics are randomly recovered. As the forward strength increases, semantics such as 'age,' 'ethnicity,' and 'glasses' are sequentially lost, and in the recovered images, these semantics are assigned random values.
    }
\label{ana1}
\end{figure*}

As the example illustrated in Figure~\ref{ana1}, the forward diffusion process leads to a gradual loss in semantics. During the experiment, we only encode the value of target semantics to the text embedding for the reverse process. Initially, at lower forward strengths (e.g., 0.15), the model successfully preserves the target semantic "smile" while keeping other attributes such as "ethnicity" unchanged. However, as the forward strength increases to 0.55 and beyond, critical semantics like "ethnicity," "glasses," and "age" are sequentially lost and replaced with random values in the reconstructed images. This progression demonstrates the trade-off in balancing required in controlling forward strength to prevent unintended alterations in semantics irrelevant to editing tasks, .

\begin{figure*}[htp] 
    \centering
    
    \includegraphics[width=1\textwidth]{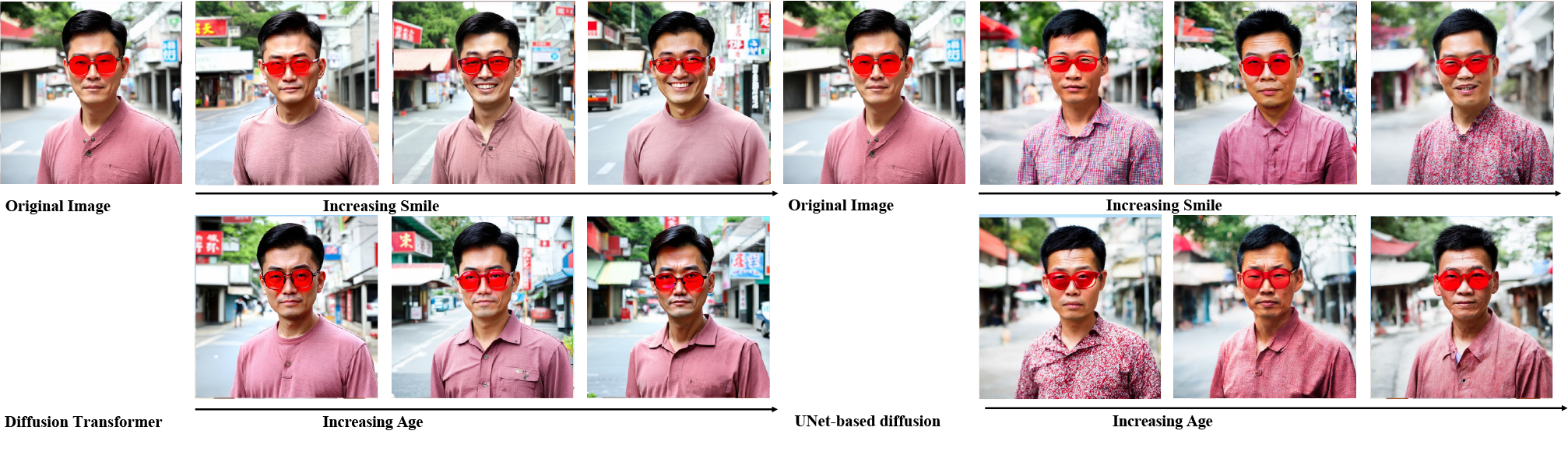} 
    \caption{Comparison of samples generated by diffusion transformer and traditional UNet-based diffusion model. Transformer-based diffusion model can achieve more precise and fine-grained image editing on target semantics.
    }
\label{sd2sd3}
\end{figure*}

\subsection{Ablation Study}
\textbf{Diffusion transformer has a better disentangled latent space.} As illustrated in Figure~\ref{sd2sd3}, the comparison between the diffusion transformer and the traditional UNet-based diffusion model clearly demonstrates the superior performance of the transformer-based approach in achieving precise and fine-grained image editing. This advantage stems from the disentanglement property inherent in the diffusion transformer. In our experiments, we applied identical text-side manipulations to the input text embeddings for both models. When increasing the "smile" attribute, the transformer model consistently produces more precise and controllable modifications, whereas the UNet-based model introduces noticeable distortions in areas unrelated to the editing task, such as clothing, hair shape, and facial structure. Similarly, when editing the "age" attribute, the diffusion transformer preserves the integrity of other facial features, resulting in more realistic outcomes compared to the UNet-based model, which struggles to isolate changes to the target semantic. This comparison highlights the advanced capabilities of transformer-based diffusion models in managing complex, fine-grained edits while maintaining overall image quality.

\textbf{Manipulating pooled text embedding as well can boost the fine-grained editing performance.} Figure~\ref{fig:ablation_pool} illustrates the impact of manipulating different text embeddings in text-guided image editing. When both text embedding and the pooled text embeddings are manipulated, the editing results are more precise and controlled, as demonstrated in the transition from a black suit to a red suit. In contrast, editing only the token embeddings results in less precise modifications, often failing to achieve the intended effect comprehensively. This highlights the importance of incorporating pooled text embeddings for achieving more refined and accurate edits.

\begin{figure}
    \centering
    \includegraphics[width=\linewidth]{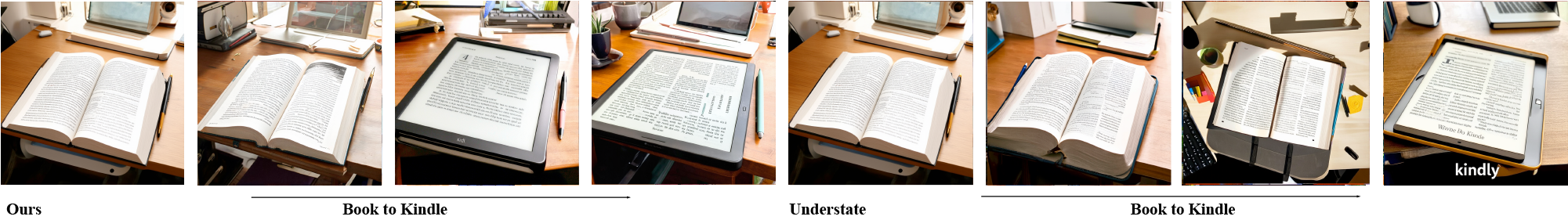}
    \caption{Highlight editing targets in the CSDS can help manipulate image embedding more effectively.}
    \label{fig:csds add}
\end{figure}

\begin{figure}
    \centering
    \includegraphics[width=1\linewidth]{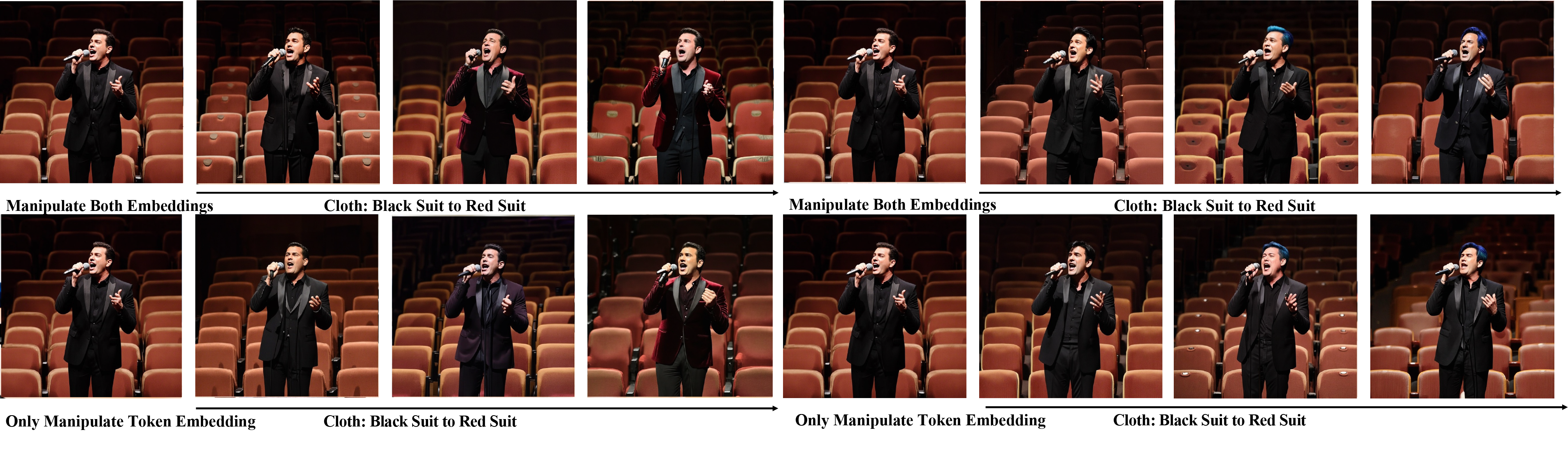}
    \caption{Manipulate both text embedding and the pooled text embedding provides more fine-grained image editing results.}
    \label{fig:ablation_pool}
\end{figure}

\textbf{Using text embeddings encoded from the text prompt of the editing target can more guide the score-distillation-sampling to modify image embedding for effectively.} Figure~\ref{fig:csds add} demonstrates that using text embeddings encoded directly from the entire text input for editing, rather than embeddings encoded specifically from the editing target, leads to less effective image manipulation. This highlights the importance of precisely targeting embeddings to maintain the quality and accuracy of image edits.

\section{Related Work}

\subsection{Diffusion Model}
In recent years, diffusion models have solidified their position as leading techniques in image generation, particularly for producing high-quality and diverse visuals~\cite{ho2020denoising, rombach2022high, esser2024scaling}. These models traditionally introduce noise to images in a forward process and utilize a U-Net architecture~\cite{ronneberger2015u} to predict the denoised outputs. Text-to-image diffusion models have particularly stood out, surpassing traditional generative models like GANs~\cite{karras2019style} in both image quality and zero-shot synthesis capabilities~\cite{rombach2022high, ramesh2022hierarchical}. While U-Net has been the standard architecture for denoising, recent research has shifted towards transformer-based diffusion models~\cite{peebles2023scalable, esser2024scaling}, which have demonstrated superior performance in generating high-fidelity and diverse images. Despite the successes of diffusion transformers, the precise mechanisms by which input conditions influence the generation process remain largely unexplored.

\subsection{Semantic Disentanglement in Generative Model}
Semantic disentanglement is an essential property in previous generative models, like Generative Adversarial Networks~\cite{goodfellow2020generative, karras2019style} and $\beta$-Variational Autoencoder~\cite{burgess2018understanding}, enabling precise control over how specific semantics are represented in the generated samples. With a generative model that has a well-disentangled latent space, we can modify the values of specific semantics in the latent space without affecting others, thereby achieving fine-grained and precise image editing~\cite{shen2020interpreting}. 
By adjusting specific directions in this latent space, it is possible to modify only the targeted semantics~\cite{harkonen2020ganspace}. While these models exhibit decent results in continuously and precisely modifying specific semantics, their generated samples are restricted in the distribution of their training datasets, and thus lack zero-shot generating ability~\cite{dhariwal2021diffusion}.

On the other hand, diffusion models~\cite{ho2020denoising}, especially text-to-image diffusion models~\cite{saharia2022photorealistic, rombach2022high}, are remarkable in synthesizing diverse and high-fidelity samples~\cite{ho2022cascaded} and composing unseen images~\cite{liu2022compositional}. ~\cite{hertz2022prompt, kwon2022diffusion, wu2023uncovering} have demonstrated that diffusion models possess a semantic latent space with word-to-semantic mappings, this property has been utilized for attention-map-based image editing and generation~\cite{cao2023masactrl,chefer2023attend,chen2024training}. Other related works~\cite{wu2023not,wang2023stylediffusion} have considered the disentanglement in the latent space of the text embedding space, or explored the feasibility of seeking disentangled latent space~\cite{lu2024hierarchical, everaert2023diffusion, everaert2024exploiting,hahm2024isometric}. However, the editing directions for precise and fine-grained changing desired semantics are not explicit, posing a non-trivial challenge in manipulating semantics in the latent space.

\subsection{Image Editing with Text-to-Image Diffusion Model}
Recently, the capability of image editing of UNet-based Text-to-Image (T2I) diffusion models has been widely studied. Inspired by the semantic meaning of image-text attention maps, ~\cite{hertz2022prompt} demonstrates a word-to-semantic relationship in UNet-based T2I diffusion models, which motivates a series of image-editing methods based on manipulating attention maps~\cite{brooks2023instructpix2pix,cao2023masactrl,chefer2023attend}. Due to the entangled latent space of UNet-based diffusion model, these approaches often can only achieve coarse grained editing. ~\cite{kawar2023imagic,wu2023uncovering} proposes to optimize image embedding to achieve disentangled editing, however, these approaches are not zero-shot, and require extra fine-tuning on model parameters. To alleviate the entangled correlation between text tokens and image semantics, another type of methods~\cite{hertz2023delta,nam2024contrastive} consider utilize score-distillation-sampling-based methods to iteratively update the noised image latents for precisely editing target items. However, while these works have improved the ability of T2I diffusion models to follow editing instructions, leveraging them for text-based fine-grained image editing remains challenging. In this paper, we have verified the disentanglement in the representation space of the diffusion transformer, and thus enable fine-grained image editing via latent space manipulation.

\section{Conclusion}
In this work, we explored the semantic disentanglement properties of the latent space of Text-to-Image diffusion transformers. Through extensive empirical analysis, we identified that the latent space of text-to-image DiTs is inherently disentangled, enabling precise and fine-grained control over image semantics. Based on these insights, we proposed a novel Extract-Manipulate-Sample (EMS) framework that leverages this disentanglement property for zero-shot fine-grained image editing. Our method demonstrates significant advancements over existing approaches, providing a robust and efficient solution for precise and fine-grained image editing without additional training. Furthermore, we introduced a benchmark and a metric to quantitatively assess the disentanglement properties of generative models. We believe our work will inspire further research and serve as a valuable resource for advancing studies in generative models.

\clearpage
\appendix


\clearpage


\begin{thebibliography}{45}
	\providecommand{\natexlab}[1]{#1}
	\providecommand{\url}[1]{\texttt{#1}}
	\expandafter\ifx\csname urlstyle\endcsname\relax
	\providecommand{\doi}[1]{doi: #1}\else
	\providecommand{\doi}{doi: \begingroup \urlstyle{rm}\Url}\fi
	
	\bibitem[Achiam et~al.(2023)Achiam, Adler, Agarwal, Ahmad, Akkaya, Aleman, Almeida, Altenschmidt, Altman, Anadkat, et~al.]{achiam2023gpt}
	Josh Achiam, Steven Adler, Sandhini Agarwal, Lama Ahmad, Ilge Akkaya, Florencia~Leoni Aleman, Diogo Almeida, Janko Altenschmidt, Sam Altman, Shyamal Anadkat, et~al.
	\newblock Gpt-4 technical report.
	\newblock \emph{arXiv preprint arXiv:2303.08774}, 2023.
	
	\bibitem[Brooks et~al.(2023)Brooks, Holynski, and Efros]{brooks2023instructpix2pix}
	Tim Brooks, Aleksander Holynski, and Alexei~A Efros.
	\newblock Instructpix2pix: Learning to follow image editing instructions.
	\newblock In \emph{Proceedings of the IEEE/CVF Conference on Computer Vision and Pattern Recognition}, pp.\  18392--18402, 2023.
	
	\bibitem[Burgess et~al.(2018)Burgess, Higgins, Pal, Matthey, Watters, Desjardins, and Lerchner]{burgess2018understanding}
	Christopher~P Burgess, Irina Higgins, Arka Pal, Loic Matthey, Nick Watters, Guillaume Desjardins, and Alexander Lerchner.
	\newblock Understanding disentangling in beta-vae.
	\newblock \emph{arXiv preprint arXiv:1804.03599}, 2018.
	
	\bibitem[Cao et~al.(2023)Cao, Wang, Qi, Shan, Qie, and Zheng]{cao2023masactrl}
	Mingdeng Cao, Xintao Wang, Zhongang Qi, Ying Shan, Xiaohu Qie, and Yinqiang Zheng.
	\newblock Masactrl: Tuning-free mutual self-attention control for consistent image synthesis and editing.
	\newblock In \emph{Proceedings of the IEEE/CVF International Conference on Computer Vision}, pp.\  22560--22570, 2023.
	
	\bibitem[Chefer et~al.(2023)Chefer, Alaluf, Vinker, Wolf, and Cohen-Or]{chefer2023attend}
	Hila Chefer, Yuval Alaluf, Yael Vinker, Lior Wolf, and Daniel Cohen-Or.
	\newblock Attend-and-excite: Attention-based semantic guidance for text-to-image diffusion models.
	\newblock \emph{ACM Transactions on Graphics (TOG)}, 42\penalty0 (4):\penalty0 1--10, 2023.
	
	\bibitem[Chen et~al.(2024)Chen, Laina, and Vedaldi]{chen2024training}
	Minghao Chen, Iro Laina, and Andrea Vedaldi.
	\newblock Training-free layout control with cross-attention guidance.
	\newblock In \emph{Proceedings of the IEEE/CVF Winter Conference on Applications of Computer Vision}, pp.\  5343--5353, 2024.
	
	\bibitem[Clark(2019)]{clark2019does}
	Kevin Clark.
	\newblock What does bert look at? an analysis of bert’s attention.
	\newblock \emph{arXiv preprint arXiv:1906.04341}, 2019.
	
	\bibitem[Couairon et~al.(2022)Couairon, Verbeek, Schwenk, and Cord]{couairon2022diffedit}
	Guillaume Couairon, Jakob Verbeek, Holger Schwenk, and Matthieu Cord.
	\newblock Diffedit: Diffusion-based semantic image editing with mask guidance.
	\newblock \emph{arXiv preprint arXiv:2210.11427}, 2022.
	
	\bibitem[Dhariwal \& Nichol(2021)Dhariwal and Nichol]{dhariwal2021diffusion}
	Prafulla Dhariwal and Alexander Nichol.
	\newblock Diffusion models beat gans on image synthesis.
	\newblock \emph{Advances in neural information processing systems}, 34:\penalty0 8780--8794, 2021.
	
	\bibitem[Esser et~al.(2024)Esser, Kulal, Blattmann, Entezari, M{\"u}ller, Saini, Levi, Lorenz, Sauer, Boesel, et~al.]{esser2024scaling}
	Patrick Esser, Sumith Kulal, Andreas Blattmann, Rahim Entezari, Jonas M{\"u}ller, Harry Saini, Yam Levi, Dominik Lorenz, Axel Sauer, Frederic Boesel, et~al.
	\newblock Scaling rectified flow transformers for high-resolution image synthesis.
	\newblock In \emph{Forty-first International Conference on Machine Learning}, 2024.
	
	\bibitem[Everaert et~al.(2023)Everaert, Bocchio, Arpa, S{\"u}sstrunk, and Achanta]{everaert2023diffusion}
	Martin~Nicolas Everaert, Marco Bocchio, Sami Arpa, Sabine S{\"u}sstrunk, and Radhakrishna Achanta.
	\newblock Diffusion in style.
	\newblock In \emph{Proceedings of the IEEE/CVF International Conference on Computer Vision}, pp.\  2251--2261, 2023.
	
	\bibitem[Everaert et~al.(2024)Everaert, Fitsios, Bocchio, Arpa, S{\"u}sstrunk, and Achanta]{everaert2024exploiting}
	Martin~Nicolas Everaert, Athanasios Fitsios, Marco Bocchio, Sami Arpa, Sabine S{\"u}sstrunk, and Radhakrishna Achanta.
	\newblock Exploiting the signal-leak bias in diffusion models.
	\newblock In \emph{Proceedings of the IEEE/CVF Winter Conference on Applications of Computer Vision}, pp.\  4025--4034, 2024.
	
	\bibitem[Goodfellow et~al.(2020)Goodfellow, Pouget-Abadie, Mirza, Xu, Warde-Farley, Ozair, Courville, and Bengio]{goodfellow2020generative}
	Ian Goodfellow, Jean Pouget-Abadie, Mehdi Mirza, Bing Xu, David Warde-Farley, Sherjil Ozair, Aaron Courville, and Yoshua Bengio.
	\newblock Generative adversarial networks.
	\newblock \emph{Communications of the ACM}, 63\penalty0 (11):\penalty0 139--144, 2020.
	
	\bibitem[Hahm et~al.(2024)Hahm, Lee, Kim, and Lee]{hahm2024isometric}
	Jaehoon Hahm, Junho Lee, Sunghyun Kim, and Joonseok Lee.
	\newblock Isometric representation learning for disentangled latent space of diffusion models.
	\newblock \emph{arXiv preprint arXiv:2407.11451}, 2024.
	
	\bibitem[H{\"a}rk{\"o}nen et~al.(2020)H{\"a}rk{\"o}nen, Hertzmann, Lehtinen, and Paris]{harkonen2020ganspace}
	Erik H{\"a}rk{\"o}nen, Aaron Hertzmann, Jaakko Lehtinen, and Sylvain Paris.
	\newblock Ganspace: Discovering interpretable gan controls.
	\newblock \emph{Advances in neural information processing systems}, 33:\penalty0 9841--9850, 2020.
	
	\bibitem[Hertz et~al.(2022)Hertz, Mokady, Tenenbaum, Aberman, Pritch, and Cohen-Or]{hertz2022prompt}
	Amir Hertz, Ron Mokady, Jay Tenenbaum, Kfir Aberman, Yael Pritch, and Daniel Cohen-Or.
	\newblock Prompt-to-prompt image editing with cross attention control.
	\newblock \emph{arXiv preprint arXiv:2208.01626}, 2022.
	
	\bibitem[Hertz et~al.(2023)Hertz, Aberman, and Cohen-Or]{hertz2023delta}
	Amir Hertz, Kfir Aberman, and Daniel Cohen-Or.
	\newblock Delta denoising score.
	\newblock In \emph{Proceedings of the IEEE/CVF International Conference on Computer Vision}, pp.\  2328--2337, 2023.
	
	\bibitem[Higgins et~al.(2018)Higgins, Amos, Pfau, Racaniere, Matthey, Rezende, and Lerchner]{higgins2018towards}
	Irina Higgins, David Amos, David Pfau, Sebastien Racaniere, Loic Matthey, Danilo Rezende, and Alexander Lerchner.
	\newblock Towards a definition of disentangled representations.
	\newblock \emph{arXiv preprint arXiv:1812.02230}, 2018.
	
	\bibitem[Ho et~al.(2020)Ho, Jain, and Abbeel]{ho2020denoising}
	Jonathan Ho, Ajay Jain, and Pieter Abbeel.
	\newblock Denoising diffusion probabilistic models.
	\newblock \emph{Advances in neural information processing systems}, 33:\penalty0 6840--6851, 2020.
	
	\bibitem[Ho et~al.(2022)Ho, Saharia, Chan, Fleet, Norouzi, and Salimans]{ho2022cascaded}
	Jonathan Ho, Chitwan Saharia, William Chan, David~J Fleet, Mohammad Norouzi, and Tim Salimans.
	\newblock Cascaded diffusion models for high fidelity image generation.
	\newblock \emph{Journal of Machine Learning Research}, 23\penalty0 (47):\penalty0 1--33, 2022.
	
	\bibitem[Ju et~al.(2023)Ju, Zeng, Bian, Liu, and Xu]{ju2023direct}
	Xuan Ju, Ailing Zeng, Yuxuan Bian, Shaoteng Liu, and Qiang Xu.
	\newblock Direct inversion: Boosting diffusion-based editing with 3 lines of code.
	\newblock \emph{arXiv preprint arXiv:2310.01506}, 2023.
	
	\bibitem[Karras et~al.(2019)Karras, Laine, and Aila]{karras2019style}
	Tero Karras, Samuli Laine, and Timo Aila.
	\newblock A style-based generator architecture for generative adversarial networks.
	\newblock In \emph{Proceedings of the IEEE/CVF conference on computer vision and pattern recognition}, pp.\  4401--4410, 2019.
	
	\bibitem[Kawar et~al.(2023)Kawar, Zada, Lang, Tov, Chang, Dekel, Mosseri, and Irani]{kawar2023imagic}
	Bahjat Kawar, Shiran Zada, Oran Lang, Omer Tov, Huiwen Chang, Tali Dekel, Inbar Mosseri, and Michal Irani.
	\newblock Imagic: Text-based real image editing with diffusion models.
	\newblock In \emph{Proceedings of the IEEE/CVF Conference on Computer Vision and Pattern Recognition}, pp.\  6007--6017, 2023.
	
	\bibitem[Kwon et~al.(2022)Kwon, Jeong, and Uh]{kwon2022diffusion}
	Mingi Kwon, Jaeseok Jeong, and Youngjung Uh.
	\newblock Diffusion models already have a semantic latent space.
	\newblock \emph{arXiv preprint arXiv:2210.10960}, 2022.
	
	\bibitem[Liu et~al.(2024)Liu, Wang, Cao, Jia, and Huang]{liu2024towards}
	Bingyan Liu, Chengyu Wang, Tingfeng Cao, Kui Jia, and Jun Huang.
	\newblock Towards understanding cross and self-attention in stable diffusion for text-guided image editing.
	\newblock In \emph{Proceedings of the IEEE/CVF Conference on Computer Vision and Pattern Recognition}, pp.\  7817--7826, 2024.
	
	\bibitem[Liu et~al.(2022)Liu, Li, Du, Torralba, and Tenenbaum]{liu2022compositional}
	Nan Liu, Shuang Li, Yilun Du, Antonio Torralba, and Joshua~B Tenenbaum.
	\newblock Compositional visual generation with composable diffusion models.
	\newblock In \emph{European Conference on Computer Vision}, pp.\  423--439. Springer, 2022.
	
	\bibitem[Liu et~al.(2019)Liu, Gardner, Belinkov, Peters, and Smith]{liu2019linguistic}
	Nelson~F Liu, Matt Gardner, Yonatan Belinkov, Matthew~E Peters, and Noah~A Smith.
	\newblock Linguistic knowledge and transferability of contextual representations.
	\newblock \emph{arXiv preprint arXiv:1903.08855}, 2019.
	
	\bibitem[Liu et~al.(2015)Liu, Luo, Wang, and Tang]{liu2015deep}
	Ziwei Liu, Ping Luo, Xiaogang Wang, and Xiaoou Tang.
	\newblock Deep learning face attributes in the wild.
	\newblock In \emph{Proceedings of the IEEE international conference on computer vision}, pp.\  3730--3738, 2015.
	
	\bibitem[Lu et~al.(2024)Lu, Wu, Chen, Wang, Bai, Qiao, and Liu]{lu2024hierarchical}
	Zeyu Lu, Chengyue Wu, Xinyuan Chen, Yaohui Wang, Lei Bai, Yu~Qiao, and Xihui Liu.
	\newblock Hierarchical diffusion autoencoders and disentangled image manipulation.
	\newblock In \emph{Proceedings of the IEEE/CVF Winter Conference on Applications of Computer Vision}, pp.\  5374--5383, 2024.
	
	\bibitem[Nam et~al.(2024)Nam, Kwon, Park, and Ye]{nam2024contrastive}
	Hyelin Nam, Gihyun Kwon, Geon~Yeong Park, and Jong~Chul Ye.
	\newblock Contrastive denoising score for text-guided latent diffusion image editing.
	\newblock In \emph{Proceedings of the IEEE/CVF Conference on Computer Vision and Pattern Recognition}, pp.\  9192--9201, 2024.
	
	\bibitem[Peebles \& Xie(2023)Peebles and Xie]{peebles2023scalable}
	William Peebles and Saining Xie.
	\newblock Scalable diffusion models with transformers.
	\newblock In \emph{Proceedings of the IEEE/CVF International Conference on Computer Vision}, pp.\  4195--4205, 2023.
	
	\bibitem[Poole et~al.(2022)Poole, Jain, Barron, and Mildenhall]{poole2022dreamfusion}
	Ben Poole, Ajay Jain, Jonathan~T Barron, and Ben Mildenhall.
	\newblock Dreamfusion: Text-to-3d using 2d diffusion.
	\newblock \emph{arXiv preprint arXiv:2209.14988}, 2022.
	
	\bibitem[Ramesh et~al.(2022)Ramesh, Dhariwal, Nichol, Chu, and Chen]{ramesh2022hierarchical}
	Aditya Ramesh, Prafulla Dhariwal, Alex Nichol, Casey Chu, and Mark Chen.
	\newblock Hierarchical text-conditional image generation with clip latents.
	\newblock \emph{arXiv preprint arXiv:2204.06125}, 1\penalty0 (2):\penalty0 3, 2022.
	
	\bibitem[Rombach et~al.(2022)Rombach, Blattmann, Lorenz, Esser, and Ommer]{rombach2022high}
	Robin Rombach, Andreas Blattmann, Dominik Lorenz, Patrick Esser, and Bj{\"o}rn Ommer.
	\newblock High-resolution image synthesis with latent diffusion models.
	\newblock In \emph{Proceedings of the IEEE/CVF conference on computer vision and pattern recognition}, pp.\  10684--10695, 2022.
	
	\bibitem[Ronneberger et~al.(2015)Ronneberger, Fischer, and Brox]{ronneberger2015u}
	Olaf Ronneberger, Philipp Fischer, and Thomas Brox.
	\newblock U-net: Convolutional networks for biomedical image segmentation.
	\newblock In \emph{Medical image computing and computer-assisted intervention--MICCAI 2015: 18th international conference, Munich, Germany, October 5-9, 2015, proceedings, part III 18}, pp.\  234--241. Springer, 2015.
	
	\bibitem[Saharia et~al.(2022)Saharia, Chan, Saxena, Li, Whang, Denton, Ghasemipour, Gontijo~Lopes, Karagol~Ayan, Salimans, et~al.]{saharia2022photorealistic}
	Chitwan Saharia, William Chan, Saurabh Saxena, Lala Li, Jay Whang, Emily~L Denton, Kamyar Ghasemipour, Raphael Gontijo~Lopes, Burcu Karagol~Ayan, Tim Salimans, et~al.
	\newblock Photorealistic text-to-image diffusion models with deep language understanding.
	\newblock \emph{Advances in neural information processing systems}, 35:\penalty0 36479--36494, 2022.
	
	\bibitem[Shen \& Zhou(2021)Shen and Zhou]{shen2021closed}
	Yujun Shen and Bolei Zhou.
	\newblock Closed-form factorization of latent semantics in gans.
	\newblock In \emph{Proceedings of the IEEE/CVF conference on computer vision and pattern recognition}, pp.\  1532--1540, 2021.
	
	\bibitem[Shen et~al.(2020)Shen, Gu, Tang, and Zhou]{shen2020interpreting}
	Yujun Shen, Jinjin Gu, Xiaoou Tang, and Bolei Zhou.
	\newblock Interpreting the latent space of gans for semantic face editing.
	\newblock In \emph{Proceedings of the IEEE/CVF conference on computer vision and pattern recognition}, pp.\  9243--9252, 2020.
	
	\bibitem[von Platen et~al.(2022)von Platen, Patil, Lozhkov, Cuenca, Lambert, Rasul, Davaadorj, Nair, Paul, Berman, Xu, Liu, and Wolf]{von-platen-etal-2022-diffusers}
	Patrick von Platen, Suraj Patil, Anton Lozhkov, Pedro Cuenca, Nathan Lambert, Kashif Rasul, Mishig Davaadorj, Dhruv Nair, Sayak Paul, William Berman, Yiyi Xu, Steven Liu, and Thomas Wolf.
	\newblock Diffusers: State-of-the-art diffusion models.
	\newblock \url{https://github.com/huggingface/diffusers}, 2022.
	
	\bibitem[Wang et~al.(2021)Wang, Yue, Huang, Sun, and Zhang]{wang2021self}
	Tan Wang, Zhongqi Yue, Jianqiang Huang, Qianru Sun, and Hanwang Zhang.
	\newblock Self-supervised learning disentangled group representation as feature.
	\newblock \emph{Advances in Neural Information Processing Systems}, 34:\penalty0 18225--18240, 2021.
	
	\bibitem[Wang et~al.(2023)Wang, Zhao, and Xing]{wang2023stylediffusion}
	Zhizhong Wang, Lei Zhao, and Wei Xing.
	\newblock Stylediffusion: Controllable disentangled style transfer via diffusion models.
	\newblock In \emph{Proceedings of the IEEE/CVF International Conference on Computer Vision}, pp.\  7677--7689, 2023.
	
	\bibitem[Wu et~al.(2023{\natexlab{a}})Wu, Liu, Zhao, Kale, Bui, Yu, Lin, Zhang, and Chang]{wu2023uncovering}
	Qiucheng Wu, Yujian Liu, Handong Zhao, Ajinkya Kale, Trung Bui, Tong Yu, Zhe Lin, Yang Zhang, and Shiyu Chang.
	\newblock Uncovering the disentanglement capability in text-to-image diffusion models.
	\newblock In \emph{Proceedings of the IEEE/CVF conference on computer vision and pattern recognition}, pp.\  1900--1910, 2023{\natexlab{a}}.
	
	\bibitem[Wu et~al.(2023{\natexlab{b}})Wu, Nakashima, and Garcia]{wu2023not}
	Yankun Wu, Yuta Nakashima, and Noa Garcia.
	\newblock Not only generative art: Stable diffusion for content-style disentanglement in art analysis.
	\newblock In \emph{Proceedings of the 2023 ACM International conference on multimedia retrieval}, pp.\  199--208, 2023{\natexlab{b}}.
	
	\bibitem[Yue et~al.(2024)Yue, Wang, Sun, Ji, Chang, Zhang, et~al.]{yue2024exploring}
	Zhongqi Yue, Jiankun Wang, Qianru Sun, Lei Ji, Eric~I Chang, Hanwang Zhang, et~al.
	\newblock Exploring diffusion time-steps for unsupervised representation learning.
	\newblock \emph{arXiv preprint arXiv:2401.11430}, 2024.
	
	\bibitem[Zhang et~al.(2024)Zhang, He, Kuang, Liu, Chen, Wu, Xiao, and Zhang]{zhang2024distributionally}
	Fengda Zhang, Qianpei He, Kun Kuang, Jiashuo Liu, Long Chen, Chao Wu, Jun Xiao, and Hanwang Zhang.
	\newblock Distributionally generative augmentation for fair facial attribute classification.
	\newblock In \emph{Proceedings of the IEEE/CVF Conference on Computer Vision and Pattern Recognition}, pp.\  22797--22808, 2024.
	
\end{thebibliography}
\end{document}